\documentclass{article}

\usepackage[nonatbib,preprint]{neurips_2021}

\usepackage[utf8]{inputenc} 
\usepackage[T1]{fontenc}    
\usepackage{hyperref}       
\usepackage{url}            
\usepackage{booktabs}       
\usepackage{amsfonts}       
\usepackage{nicefrac}       
\usepackage{microtype}      
\usepackage{xcolor}         
\usepackage{paralist}
\usepackage{csquotes}
\usepackage{wrapfig}
\usepackage{color}
\usepackage{colortbl}
\usepackage{color}
\usepackage{csquotes}
\usepackage{soul}
\usepackage{enumerate}
\usepackage[normalem]{ulem}
\usepackage{algorithm}
\usepackage{algorithmicx}
\usepackage{algpseudocode}
\usepackage{booktabs}
\usepackage{array}
\usepackage{rotating}
\usepackage{arydshln}
\usepackage{ragged2e}
\usepackage{wrapfig}

\newcommand\ie{\emph{i.e., }}

\usepackage{authblk}
\usepackage{amsmath}
\DeclareMathOperator{\tr}{tr}
\newcommand{\norm}[1]{\left\lVert#1\right\rVert}
\usepackage{bm}
\usepackage{algorithm}
\usepackage{graphicx}
\usepackage{multirow}
\usepackage{enumitem}
\setlist[itemize]{left=5pt, itemsep=0pt,partopsep=0pt,parsep=0pt,topsep=0pt}
\newcommand{\Lip}{\mathbb{L}}
\newcommand{\R}{{\mathbb R}}
\newcommand{\calX}{{\cal X}}
\definecolor{mygray}{gray}{.9}

\newcommand{\abs}[1]{\left\lvert#1\right\rvert}

\title{Invertible Attention}

\author{Jiajun Zha$^{1,2}$, Yiran Zhong$^{1}$, Jing Zhang$^{1}$, Richard Hartley$^{1}$, Liang Zheng$^{1}$,
\\
$^{1}$Australian National University,
$^{2}$Fudan University\\
\texttt{\{u7045661, firstname.lastname\}@anu.edu.au, zhongyiran@gmail.com}}

\begin{document}

\maketitle

\begin{abstract}
Attention has been proved to be an efficient mechanism to capture long-range dependencies. However, so far it has not been deployed in invertible networks. This is due to the fact that in order to make a network invertible, every component within the network needs to be a bijective transformation, but a normal attention block is not. In this paper, we propose invertible attention that can be plugged into existing invertible models. We mathematically and experimentally prove that the invertibility of an attention model can be achieved by carefully constraining its Lipschitz constant. We validate the invertibility of our invertible attention on image reconstruction task with 3 popular datasets: CIFAR-10, SVHN, and CelebA. We also show that our invertible attention achieves similar performance in comparison with normal non-invertible attention on dense prediction tasks.
\end{abstract}


\section{Introduction}
There is a growing interest in invertible networks, especially in generative tasks~\cite{papamakarios2019normalizing,NFReview,SRFlow}. An invertible network often refers as a bijective function whose input and output are one-to-one correlated:\begin{equation}
\textstyle{
    \mathcal{F}:\mathbb{R}^d \to \mathbb{R}^d, \bm{x}\mapsto\bm{z}; \quad \mathcal{F}^{-1}:\mathbb{R}^d \to \mathbb{R}^d, \bm{z}\mapsto\bm{x}. 
}\end{equation}
With this property, we can model generative models with exact likelihood to stabilise the training process, analyze the invariance of a network, and solve inverse problems.

However, designing an invertible network is a non-trivial task as it requires every component within the network to be invertible. For a common feedforward network, it means every weight matrix has to be non-singular and some commonly used activation functions such as ReLU \cite{relu} cannot be applied as they are not bijective. A necessary condition of an invertible network is that it requires the size of input and output for each neural block to be the same. This condition becomes a barrier to stack encoder-decoder structures in order to increase the receptive field, which we normally do in a convolutional deep network~\cite{hourglass}.

To increase the receptive field for an invertible network, numerous methods have been proposed such as partitioning dimensions~\cite{dinh2016density} (\ie squeezing the spatial dimension to the channel dimension) and stacking convolutions~\cite{truong2019generative}. Since none of them are able to directly capture long-range dependencies, current invertible networks usually have limited receptive field.

Attention~\cite{vaswani2017attention} or non-local network~\cite{Non-localNN} is a mechanism aiming to capture long-range/global dependencies. It has been widely used in both natural language processing~\cite{devlin2018bert,brown2020language} and computer vision tasks~\cite{fu2019dual,zhao2019pyramid,Wang2020Parallax}. A general form of attention can be expressed as
\begin{equation}
\label{eq:att}\textstyle{
    \bm{A}(\bm{x})_i = \frac{1}{\bm{N}(\bm{x})}\sum_{\forall j}\bm{r}(\bm{x}_i, \bm{x}_j)\bm{F}(\bm{x})_j,
}\end{equation}
where $\bm{x}, \bm{A}(\bm{x})$ are the input and output signals, respectively. $i,j$ represents all possible positions' index. $\bm{F}(\cdot)$ projects the input $\bm{x}$ to feature space and $\bm{N}(\bm{x})$ is a normalizing factor. $\bm{r}(\cdot)$ is the response function that compute the correlation between $i$ and $j$. We also define response map $\bm{R}(\bm{x})$ as the set of response values $\bm{r}(\bm{x}_i, \bm{x}_j)$ for all possible pairs of $i,j$ after normalizing. As shown in Fig. \ref{fig: basic attention}, the response map is often a square matrix. Since the computation on one position considers all possible positions, the attention module can capture all-range dependencies. 

However, an attention module is not naturally invertible. In this paper, we prove that an attention module can become invertible by carefully constraining its Lipschitz constant. Specifically, we impose mild restrictions on the response map, feature mapping $\bm{F}$ and bring in a Lipschitz constrained convolution at the last step of residual branch.
We mathematically prove, and practically validate the invertibility of our method with four kinds of attentions (\emph{i.e.,} Gaussian, Embedded Gaussian, Dot-product and Concatenation) on the image reconstruction task and demonstrate that the input images can be nearly perfectly reconstructed. 
We then show our module could be embed into existing invertible structures for generative tasks. Moreover, to analyze the effect of imposing these constraints, we compare the performance of our invertible attention and the original non-invertible attention on a dense prediction task. Results show that our invertible attention is roughly on par with the normal non-invertible attentions. 



\section{Methodology}
\label{sec: methodology}
\begin{figure}
  \centering
  \includegraphics[width=\linewidth]{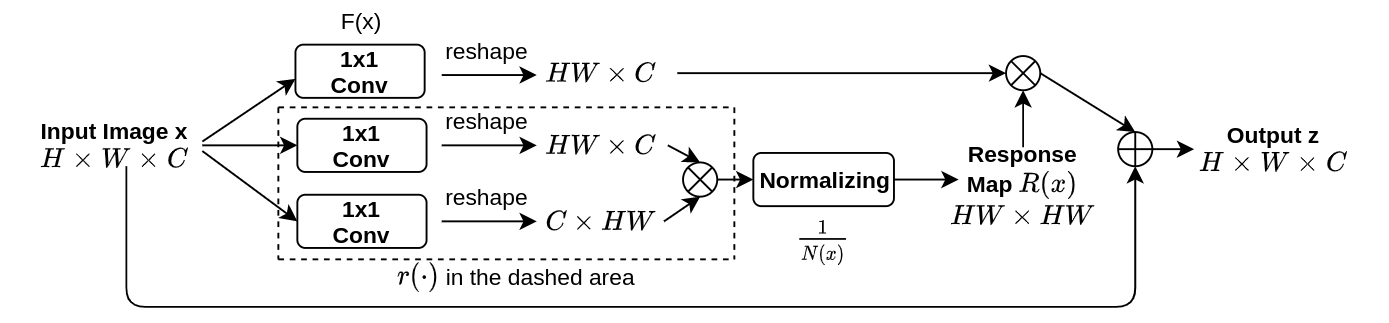}
  \caption{A Dot-product attention block. Each of the three copies of the input image goes through a different $1\times 1$ convolution to get a feature map. Two of the feature maps perform a matrix multiplication with each other, and then multiply a normalizing coefficient to get a response map. The response map then perform another matrix multiplication with the last feature map, and the product will be added back to the input.}
  \label{fig: basic attention}
\end{figure}

An attention module often presents in a residual form~\cite{Non-localNN}. Depending on the selection of response function $\bm{r}(\cdot)$, we have 4 kinds of attentions, namely Gaussian, Embedded Gaussian, Dot-product and Concatenation. Figure~\ref{fig: basic attention} illustrates a widely used Dot-product attention block. It generally consists three steps: 1) building a response map by computing the correlation between each pair of positions; 2) using the response map to perform a weighted sum over all positions by matrix multiplication with the input feature map; 3) adding the weighted feature map to the input with residual connections. 
Since the invertible residual structure and the invertible convolution are involved in our invertible attention module, for the sake of completeness, we briefly describe them first, and then introduce our invertible attention module.


\subsection{Invertible Residual Structure}
A feedforward neural network with a shortcut connection is referred as a residual structure. Specifically, given an input $\bm{x}\in \R^d$, the residual structure can be written as: $\bm{H}(\bm{x}) = \bm{x}+\bm{G}(\bm{x})$. In order to make $\bm{H}$ invertible, one sufficient condition is to constrain the Lipschitz constant $\mathbb{L}$ of $\bm{G}$~\cite{behrmann2019invertible}.

\textbf{Lemma 1:} \emph{$\bm{H}(\bm{x})$ is invertible if $\Lip(\bm{G}) = c \text{ where }c\in (0, 1)$, and $\Lip(\bm{G})$ is defined as
\begin{equation}
\label{eq:invresnet}
\textstyle{
    \quad \Lip(\bm{G}) = \sup_{\bm{x}_1 \neq \bm{x}_2} \frac{\norm{\bm{G}(\bm{x}_1) - \bm{G}(\bm{x}_2)}}{\norm{\bm{x}_1 - \bm{x}_2}}.
}\end{equation}}
The inverse of $\bm{H}$ can be computed by a fixed-point iteration algorithm~\cite{behrmann2019invertible}.
\setlength{\textfloatsep}{0.4cm}
\setlength{\floatsep}{0.2cm}
\begin{algorithm}
\caption{Iterative Inverse of Residual Block}\label{alg:IterInv}
\begin{algorithmic}[1]
\Function{INV}{$\bm{z}, \bm{G}$}\Comment{Output $\bm{z}$ of the residual block and the residual transformation $\bm{G}$}
\State $\bm{x}^0\gets \bm{z}$
\For{$i=0\dots N-1$} \Comment{$N$ is a user-specified number}
\State $\bm{x}^{i+1} \gets \bm{z} -  \bm{G}(\bm{x}^i)$
\EndFor
\State \textbf{return} $\bm{x}^N$\Comment{$\bm{x}^N$ is numerically very close to $\bm{x}$}
\EndFunction
\end{algorithmic}
\end{algorithm}

The core idea of this algorithm is numeric estimation. The convergence speed and training stability depend on two hyperparameters $\Lip(\bm{G})$, and $N$. We empirically set $\Lip(\bm{G})=0.9, N=100$. Note the sufficient conditions for invertibility here are 1) the residual structure and 2) the Lipschitz constraint on the residual branch. Since the attention is also a residual structure, a logical heuristic is to apply Lipschitz constraint on the attention residual branch to achieve invertibility. 


\subsection{Invertible Convolution}
\label{sec: inv vonc}
A convolution is a linear transformation and can be represented by
$
\bm{g}(\bm{x}) = \bm{W}\bm{x},
$
where $\bm{W}$ is the weight matrix of convolution.
For $L_2$-norm, $\Lip(\bm{g})$ is defined as the largest singular value of $\bm{W}$. It is less preferable to use the singular value decomposition (SVD) to find $\Lip(\bm{g})$ because it is time-consuming and does not take advantage of parallel computing hardware. Another algorithm that achieves the same purpose but fits parallel computing more properly is Power Iteration \cite{mises1929praktische}.
With the computed largest singular value $\sigma(\bm{W})$,
we can enforce the Lipschitz constant constraint by a simple normalization:
\begin{equation}
\textstyle{
\bm{W} = 
\begin{cases}
\frac{c\bm{W}}{\sigma(\bm{W})}\quad  \text{if } \frac{c}{\sigma(\bm{W})} < 1\\
\bm{W}  \quad\quad \text{else}
\end{cases}.
}
\end{equation}


\subsection{Invertible Attention}

The residual structure of attention provides the possibility
to invert it by constraining its Lipschitz constant. However, it is still a non-trivial task since we need to know how to constrain the
Lipschitz constant while maintaining similar capacity and
functionality as normal attentions. In the following sections, we first derive a theorem which gives a sufficient condition to invert attention, then introduce practical implementation methods.

\subsubsection{Constrain the Lipschitz constant of an attention}

As shown in Eq.~\ref{eq:att}, an attention can be seen as a matrix multiplication between a response map and a feature map. Therefore, constraining the Lipschitz constant of an attention is equivalent to calculate the Lipschitz bounds for products of matrices. Here, for simplicity, we start with $L_1$ norms and then extend it to $L_2$ norms.



\textbf{Theorem 1:}
Let $\calX$ be a normed vector space and
let $\R^{m\times n}$ represents the
set of $m \times n$ vector spaces over real number $\R$.
Let $\bm{F}: \calX \to \R^{m\times n}$, and $\bm{R}: \calX \to \R^{m\times m}$. Define $\bm{A}:\calX \to \R^{m\times n}$ by $\bm{A}(\bm{x}) = \bm{R}(\bm{x})\bm{F}(\bm{x})\text{ where } x\in \calX$~. We further assume the following properties:
\begin{compactenum}
\item $\bm{F}$\  \textit{is Lipschitz-continuous with $L_1$-Lipschitz constant}\ $c_{{F}}$.
\item   $\bm{R}$\  \textit{is Lipschitz-continuous with $L_1$-Lipschitz constant}\ $c_{{R}}$.
\item  $\norm{\bm{R}(\bm{x})}_1 \leq \mu_{{R}} \text{ for each }\bm{x}\in \calX$.
\item  $\norm{\bm{F}(\bm{x})}_1 \leq \mu_{{F}} \text{ for each }\bm{x}\in \calX$.
\end{compactenum}
Then $\bm{A}$ has a $L_1$-Lipschitz constant $\mu_{{R}}c_{F} + \mu_Fc_R$.

\textit{Proof:} For simplicity, denote $\bm{R}(\bm{x}^i)$ and $\bm{F}(\bm{x}^i)$  by $\bm{R}_i$ and $\bm{F}_i$ respectively. Using triangle inequality and the sub-multiplicative property of $L_1$ norms, we have:
\small
\begin{align*}
\|\bm{A}( \bm{x}^1) - \bm{A}( \bm{x}^2)\|_1 &= \| \bm{R}_1 \bm{F}_1 - \bm{R}_2 \bm{F}_2 \|_1 
 = \| \bm{R}_1 (\bm{F}_1 - \bm{F}_2) + (\bm{R}_1 - \bm{R}_2) \bm{F}_2 \|_1 \\
 &\le \|\bm{R}_1 (\bm{F}_1 -\bm{F}_2)\|_1 + \| (\bm{R}_1 - \bm{R}_2) F_2 \|_1 
 \le \| \bm{R}_1 \|_1 \| \bm{F}_1 - \bm{F}_2 \|_1 + \| \bm{R}_1 - \bm{R}_2 \|_1 \|\bm{F}_2\|_1 \\
 &\le \mu_R  c_F \| \bm{x}^1 -  \bm{x}^2 \|_1 + c_R \|  \bm{x}^1 -  \bm{x}^2 \|_1 \mu_F ~,
\end{align*}
\normalsize
\textbf{Extension: }

Since the $L_1$ norm and $L_2$ norm are within constant bounds of each other, a function is $L_1$ Lipschitz continuous if and only if it is $L_2$ Lipschitz continuous. Furthermore, the derivation above gives a tighter bound at the sub-multiplicative inequality step, which gives direct inspiration to our practical implementation. The proof on $L_2$-norm are given in Appendix \ref{sec: supp norm extension}.


\subsubsection{Residual Attention}
Given the proved $L_2$ Lipschitz constant $\Lip(\bm{A}) = k(\mu_Rc_F + c_R\mu_F)$, where $k$ denoting the constant bound between $L_1$ and $L_2$ norm, we add another Lipshictz constrained convolution at the end of the residual branch. So the whole attention block now is
\begin{equation}
    \bm{f}(\bm{x}) = \bm{x} + \bm{W}_L\bm{A}(\bm{x}), 
\end{equation}
where $\bm{W}_L$ is the weight matrix of the last convolution.
We set the Lipschitz constant of this convolution to be $\frac{c}{k(\mu_Rc_F + c_R\mu_F)}$, where $c\in(0,1)$. Therefore the residual branch satisfies
\begin{equation}
    \mathbb{L}(\bm{W}_L\bm{A}(\bm{x}))  \mathbb{L}(\bm{W}_L\bm{x})\mathbb{L}(\bm{A}(\bm{x})) = \frac{c}{k(\mu_Rc_F + c_R\mu_F)}k(\mu_Rc_F + c_R\mu_F) = c~.
\end{equation}
Together with Lemma 1, the whole attention block $\bm{f}(\bm{x})$ could be inverted by Algorithm \ref{alg:IterInv}.


\subsection{Practical Implementation and Trade-off}
\label{sec: practical implementation}
\begin{figure}[t]
  \centering
  \includegraphics[width=\linewidth]{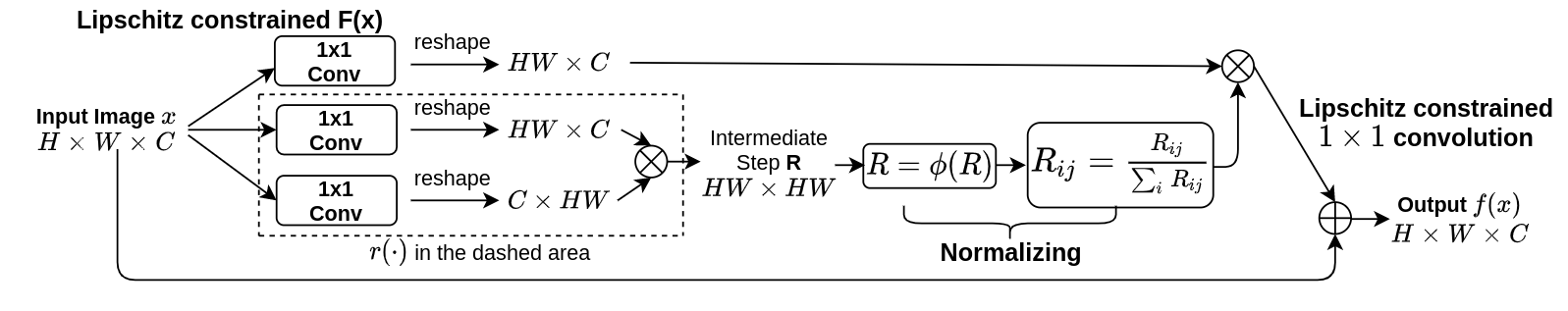}
  \caption{Process diagram of invertible Dot-product attention. To achieve invertibility, there are three places diffent from Figure \ref{fig: basic attention}. Lipschitz constrained $\bm{F}(\bm{x})$ is applied on the $1\times 1$ convolution outside the dashed area. Normalizing coefficient changes to a specific two step operation. And another Lipschitz constrained $1\times 1$ convolution is added at the last step of the residual branch.}
  \label{fig: constrained attention}
\end{figure}

As we use a chain of inequalities in the derivation, the final Lipschitz bound may still hold true even if we break some inequalities within that chain. Furthermore, if strictly following the sufficient conditions to design invertible attention, the expressive power of such model is severely limited. A trade-off is needed between invertibility and expressive power. 
We empirically give the following relaxed settings that generally perform well in our experiments in terms of both invertibility and expressive power.
\begin{itemize}
\setlength{\parskip}{4pt}
\item \scalebox{1.0}[1.0]{\textit{$\Lip (\bm{F}) = 0.9$ with respect to $L_2$ norm}} 

This corresponds to the first condition in the theorem 1. It is implemented by invertible convolution techniques introduced in Sec. \ref{sec: inv vonc}.
\item \scalebox{1.0}[1.0]{\textit{$\bm{M}_{ij} \geq 0 \text{ for all }i,j$ and $ \sum_i\bm{M}_{ij} = 1\text{ for all }j$}} 

This restriction is actually stronger than the third condition in the proof, specifically, stronger than
$
    \norm{\bm{R}(\bm{x})}_1 = 1~.
$
We choose this implementation as it keeps the framework almost the same as vanilla attention, 
and does not introduce a significant increase on computation cost. 
This technique is implemented by first wrapping the response value with an activation function $\phi$ satisfying $\phi(x) \geq 0 \text{ for all }x\in\R$, and then change the normalizing coefficient to 
$
    \bm{N}(\bm{x}) = \sum_{i=1}^m\bm{r}(\bm{x}_i,\bm{x}_j)~.
$



\item \scalebox{1.0}[1.0]{\textit{$\Lip(\bm{W}_L\bm{x}) = 0.9$ with respect to $L_2$ norm}} 

$\bm{W}_L$ is the weight matrix of the last $1\times 1$ convolution on the residual branch.
This could be seen as the scaling factor we add at the end of the residual branch. We can implement this condition in the same way as $\Lip (\bm{F}) = 0.9$~.
\end{itemize}

\textbf{Explanation: } In practice, $\bm{R}(\bm{x})$ is computed by different approaches across different types of attention, 
and there is no easy way to directly confine its Lipschitz constant. 
So the second condition is not implemented, 
and our experiments show that it still performs generally well. Besides, we empirically find that 
the numeric value of a normally behaved neural network falls within a small range around zero, so the fourth condition naturally holds. We do not use any method to confine it here.  

Each of the practical implementation techniques above is mild and easy to realize. 
They bring no significant change to the existing framework of attention. 
For better understanding, we demonstrate the process of invertible Dot-product attention in Fig.~\ref{fig: constrained attention}.
Following \cite{Non-localNN}, 
we also present four kinds of attention in both non-invertible and invertible form in Table \ref{attention_table}.

\begin{table}
\small
  \caption{Comparison of non-invertible \& invertible attention. According to the ways to calculate the response (Column 4), four types of attention are presented. For the proposed invertible attention, we use \textbf{bold} words to highlight the differences from non-invertible attention.}
  \label{attention_table}
  \centering
  \setlength{\tabcolsep}{3.4mm}
  \begin{tabular}{llccc}
    \toprule
     Invertibility & Type  & \textbf{Focus} $\bm{F}(\bm{x}_j)$   & \textbf{Response $\bm{r}(\bm{x}_i. \bm{x}_j)$} & \textbf{Norm. Coef. $\bm{N}(\bm{x})$}\\
    \midrule
\multirow{4}{*}{Non-invertible} & Gaus.          & \multirow{4}{*}{$1\times1$ Conv} & $e^{\bm{x}_i^\top\bm{x}_j}$                             & \multirow{2}{*}{$\sum_{\forall {j}}\bm{r}(\bm{x}_i,\bm{x}_j)$} \\ 
& Embed. &                                  & $e^{(\bm{W}_1\bm{x}_i)^\top(\bm{W}_2\bm{x}_j)}$         &                                                              \\ \cline{5-5} 
& Dot.  &                                  & $(\bm{W}_1\bm{x}_i)^\top(\bm{W}_2\bm{x}_j)$             & \multirow{2}{*}{$\sum_{\forall {j}}1$}                         \\
& Concat. &                                  & $\bm{W}_3[\bm{W}_1\bm{x}_i, \bm{W}_2\bm{x}_j]$&\\
    \midrule 
\multirow{4}{*}{Invertible} &   Gaus.        & \multirow{4}{*}{\shortstack[c]{\textbf{Lipschitz} \\ \textbf{constrained} \\ $1\times1$ Conv}} & $e^{\bm{x}_i^\top\bm{x}_j}$                             & \multirow{4}{*}{$\sum_{\forall \textbf{i}}\bm{r}(\bm{x}_i,\bm{x}_j)$} \\
& Embed. &                                                        & $e^{(\bm{W}_1\bm{x}_i)^\top(\bm{W}_2\bm{x}_j)}$         &                                                              \\
& Dot.   &                                                        & $\bm{\phi}((\bm{W}_1\bm{x}_i)^\top(\bm{W}_2\bm{x}_j))$       &                                                              \\
& Concat.  &                                                        & $\bm{\phi}(\bm{W}_3[\bm{W}_1\bm{x}_i, \bm{W}_2\bm{x}_j])$ &\\
    \bottomrule
  \end{tabular}
\end{table}


\section{Related Work}
\textbf{Attention:} It is a mechanism in deep learning that mimics the human cognitive process of concentrating on a few parts of an object while ignoring others. This mechanism was firstly introduced in natural language processing (NLP) tasks in~\cite{cho2014learning,sutskever2014sequence} and applied to computer vision later~\cite{xu2015show}. The core idea is to assign different weights to different positions of a feature to improve system performance. The weighting schemes can be applied to pixel locations (spatial attention)~\cite{Non-localNN}, pixel channels (channel attention)~\cite{wang2017residual} of an image, sentence tokens~\cite{vaswani2017attention}, and graph nodes~\cite{velivckovic2017graph}. 
One of the advantages of attention is that it allows the network to achieve a global receptive field at a low cost. So it is now widely used for long-range dependence capturing in semantic segmentation~\cite{Fu_2019_CVPR}, action recognition~\cite{xie_2018_memory}, depth estimation~\cite{Huyun_2020_ECCV} and optical flow estimation~\cite{zhai_2020_optical}. In this paper, we study spatial attention in invertible networks. 

\textbf{Receptive field in invertible networks:}
A necessary condition for a network to be invertible is that the size of its input and output should be the same. 
This condition limits the receptive field in an invertible network. Generally, 
there are two strategies to increase the receptive field in invertible networks:
partitioning dimensions and stacking convolutions. The former often copes with a multi-scale architecture and it squeezes the spatial dimension into the feature channel,
\ie reshape $C\times H\times W$ to $4C\times H/2 \times W/2$, and processes it with $1\times 1$ convolutions~\cite{dinh2016density, dinh2015nice}. 
The later is similar to conventional CNN networks where the receptive field grows gradually by stacking $n\times n$ invertible convolutions~\cite{InvConvFlow,truong2019generative,hoogeboom2019emerging}. 
However, all of these methods must be repeated multiple times to increase the receptive field and they are not efficient to achieve global receptive fields. 
Our method, on the other hand, allows invertible networks to achieve global receptive field in one block.

\section{Experiment}
In this section, we first validate the inveribility of our module on three datasets. And for each dataset, we test four types of invertible attention including Gaussian, Embedded Gaussian, Dot-product, and Concatenation. We then embed our invertible attention module into i-ResNet~\cite{behrmann2019invertible}, a recent Normalizing Flow architecture to test its capacity on image generation tasks. We also compare the expressive power of our invertible attention with non-invertible attention on the camouflaged object detection task.

\subsection{Validating Invertibility}
\textbf{Implementation:}
Our network consists of a squeeze layer and an invertible attention module. The squeeze layer is used to reduce the memory consumption. The Lipschitz constant is set to $0.9$ during training and the inverse iteration number is set to be $100$ in testing. More details are provided in Appendix \ref{sec: supp valid inv}.
We evaluate our invertible attention module on CIFAR-10 \cite{krizhevsky2009learning}, SVHN \cite{netzer2011reading}, and CelebA \cite{liu2015faceattributes} datasets under the resolution of $32\times 32$. For each dataset, we randomly select 1,000 images for training and another 1,000 for testing. For each experiment, we train small images for 100, large images for 1 epoch, and report the average scores of 3 times repeated experiments. We use MSE loss to train our network and use Mean Squared Error (MSE) and structural similarity (SSIM) as our quantitative metrics\footnote{Note that we do not use the peak signal-to-noise ratio (PSNR) metric because it will give infinity for perfect reconstruction results, which are often the case in our system.}. We also introduce the Valid score (V-score) to measure the success rate of image reconstruction. It is defined as the ratio of images that are reconstructed with MSE less than 10. We also validate it under a higher resolution of $218\times 178$ on the CelebA dataset.

\textbf{Quantitative results:} We evaluate $32\times 32$ color image reconstruction results in Table \ref{tab: QuantiRecon}, where the four types of invertible attention are compared. It shows that the designed attention module is mostly invertible for Gaussian, Embed Gaussian and Concatenation, with V-score around 90\%. Note that the MSE is computed on the RGB range of $0-255$, so even the largest error of 461.488 is still relatively small comparing to the scale of a color pixel. 

\begin{figure}[t]
    \centering
    \includegraphics[width=\linewidth]{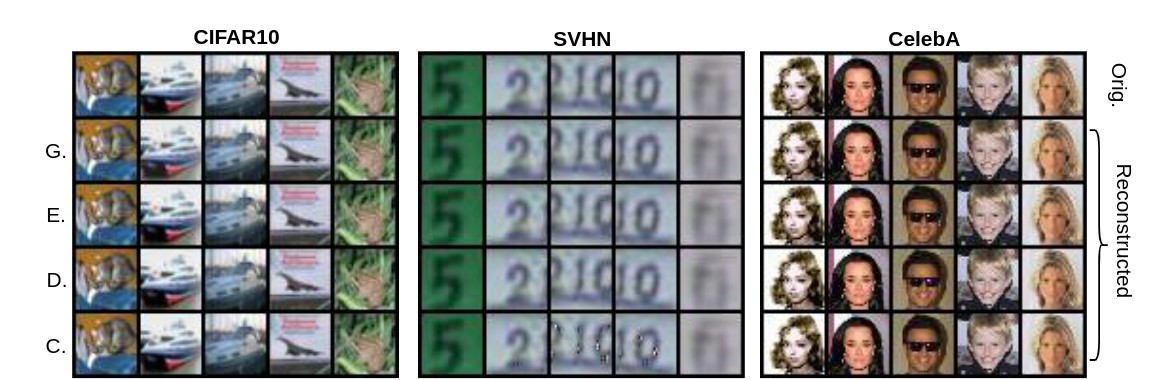}
    \caption{Reconstructed $32\times32\times3$ images by four types of invertible attention, \emph{i.e.,} Gaussian (G.), Embedded Gaussian (E.), Dot-product (D.) and Concatenation (C.) on CIFAR-10, SVHN and CelebA. Top row shows original images, and the rest rows present reconstructed ones. We clearly see that the reconstructed images look very much the same with original ones.}
    \label{fig:recon}
\end{figure}

\begin{figure}[t]
    \centering
    \includegraphics[width=\linewidth]{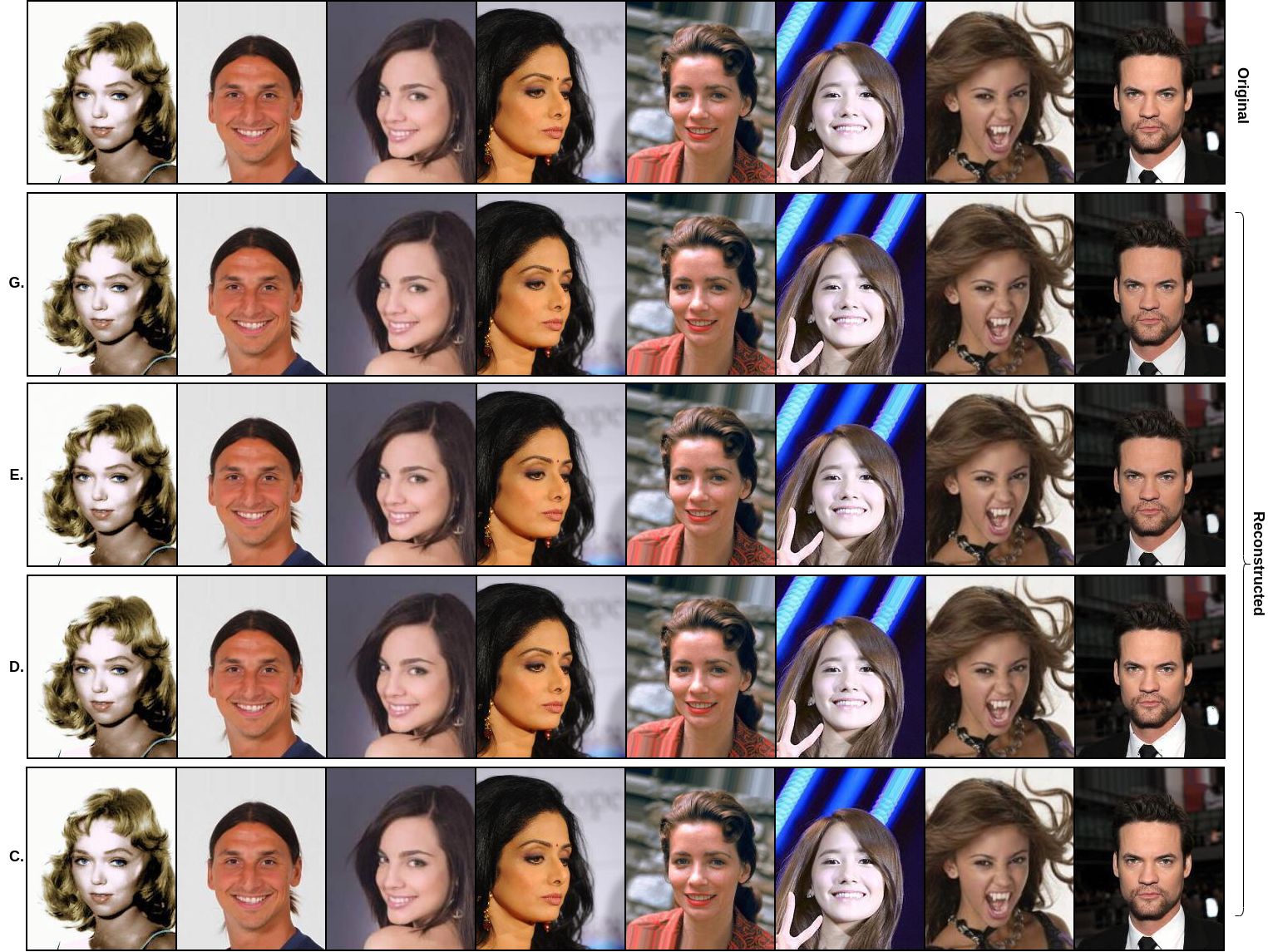}
    \vspace{-2mm}
    \caption{Images of size $218\times178\times 3$ reconstructed by four types of invertible attention: Gaussian, Embedded Gaussian, Dot-product and Concatenation on CelebA aligned face images. Our method is thus shown to be effective for large-sized image reconstruction.}
    \label{fig: large recon epoch1}
\end{figure}

\textbf{Qualitative results:} We provide qualitative results in Fig.~\ref{fig:recon}, where the reconstructed images look almost the same with the original images. 
Further, in Fig.~\ref{fig: large recon epoch1}, we provide reconstruction results of images of size $218\times178$, and again the reconstructed images are very close to the original ones. It is worth noting that being able to reconstructing larger-sized images is non-trivial for invertible networks (due to the high GPU memory consumption) and our module can handle it effectively. 

\begin{table}[t]
\small
\centering
\caption{Quantitative evaluation of reconstruction quality over four types of invertible attention. Since our model may fail for some images, these extreme values will significantly raise the MSE value. We use V-score for better measuring the success rate of reconstruction averaged on each image. }
\setlength{\tabcolsep}{1mm}
\begin{tabular}{lccccccccc}
\midrule
\multirow{2}{*}{Attention Method} & \multicolumn{3}{c}{CIFAR-10} & \multicolumn{3}{c}{SVHN} & \multicolumn{3}{c}{CelebA} \\ 
                      & MSE$\downarrow$            & SSIM$\uparrow$   & V-score$\uparrow$     & MSE$\downarrow$          & SSIM$\uparrow$    & V-score$\uparrow$   & MSE$\downarrow$           & SSIM$\uparrow$   & V-score$\uparrow$     \\ \midrule
Gaussian                  & 84.872         & 0.988    &$85.625\%$   & 0.321        & 0.999  &$99.995\%$    & 55.236        & 0.992   &$86.979\%$    \\ 
Embed. Gaussian              & 6.911          & 0.999   &$98.333\%$    & 0.003        & 0.999  &$100.000\%$    & 29.026        & 0.996   &$93.958\%$    \\
Dot-product              & 169.032        & 0.977  &$37.812\%$     & 508.016      & 0.892  &$44.583\%$    & 461.488       & 0.936    &$37.812\%$   \\ 
Concatenation               & 0.706          & 0.999  &$98.437\%$     & 171.755      & 0.994  &$97.500\%$    & 5.064         & 0.998   &$88.021\%$    \\ \bottomrule
\end{tabular}
\label{tab: QuantiRecon}
\end{table}

\subsection{Generative Modelling}
\textbf{Implementation:} We embed the invertible attention module to i-ResNet~\cite{behrmann2019invertible} and demonstrate its image generation ability on the CIFAR-10~\cite{krizhevsky2009learning} dataset. To apply our module to the image generative task, an efficient algorithm to compute the log determinant of its Jacobian matrix is needed. We adapt the one proposed in i-ResNet~\cite{behrmann2019invertible} to our model. Details are provided in \ref{sec: supp generative modelling}.

\begin{figure}[t]
    \centering
    \includegraphics[width=\linewidth]{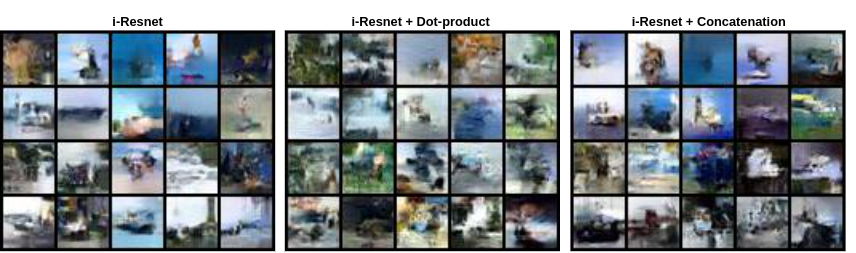}
    \vspace{-2mm}
    \caption{Sample images generated by vanilla i-ResNet (left),   invertible Dot-product attention embedded in i-ResNet (middle) and invertible Concatenation attention embedded in i-ResNet (right). Training is performed on CIFAR-10.} 
    \label{fig: iResnet}
\end{figure}

\begin{table}[h]
\label{tab: Recon}
\scriptsize
\centering
\caption{Comparison on density estimation with existing normalizing flow models on CIFAR-10. ``i-ResNet+D.'' and ``i-ResNet+C.'' denote i-ResNet embedded with invertible Dot-product attention and invertible Concatenation attention, respectively. We use bits/dim as the evaluation metric.}
\setlength{\tabcolsep}{1.6mm}
\begin{tabular}{ccccccccc}
\toprule
Model               & MADE~\cite{pmlr-v37-germain15} & MAF~\cite{MAF}  & RealNVP~\cite{dinh2016density} & Glow~\cite{kingma2018glow} & FFJORD~\cite{grathwohl2018scalable} & i-ResNet~\cite{behrmann2019invertible} & i-ResNet + D. & i-ResNet + C. \\ \midrule
bits/dim$\downarrow$ & $5.67$ & $4.31$ & $3.49$    & $3.35$ & $3.40$   & $3.45$     & $3.65$                   & $3.39$                   \\ \bottomrule
\end{tabular}
\vspace{3mm}
\end{table}

\textbf{Results:} We train the original i-ResNet, i-ResNet with invertible Concatenation attention, and i-ResNet with invertible Dot-product attention under the same setting as \cite{behrmann2019invertible} on CIFAR-10. Invertible Gaussian and embedded Gaussian attention are not stable in this experiment due to their intrinsic numeric issues, details in Appendix \ref{sec: supp gaussian issue}.
 
Some generated samples are shown in Fig.~\ref{fig: iResnet}. Quantitative results measuring the model capacity with respect to bits/dim metric~\cite{MAF} are shown in Table~\ref{tab: Recon}. Comparing with original i-ResNet, our method improves by $1.8\%$ through adding only one invertible attention block. 

\begin{figure}[t]
    \centering
    \begin{tabular}{c@{ }}
    \includegraphics[width=0.99\linewidth]{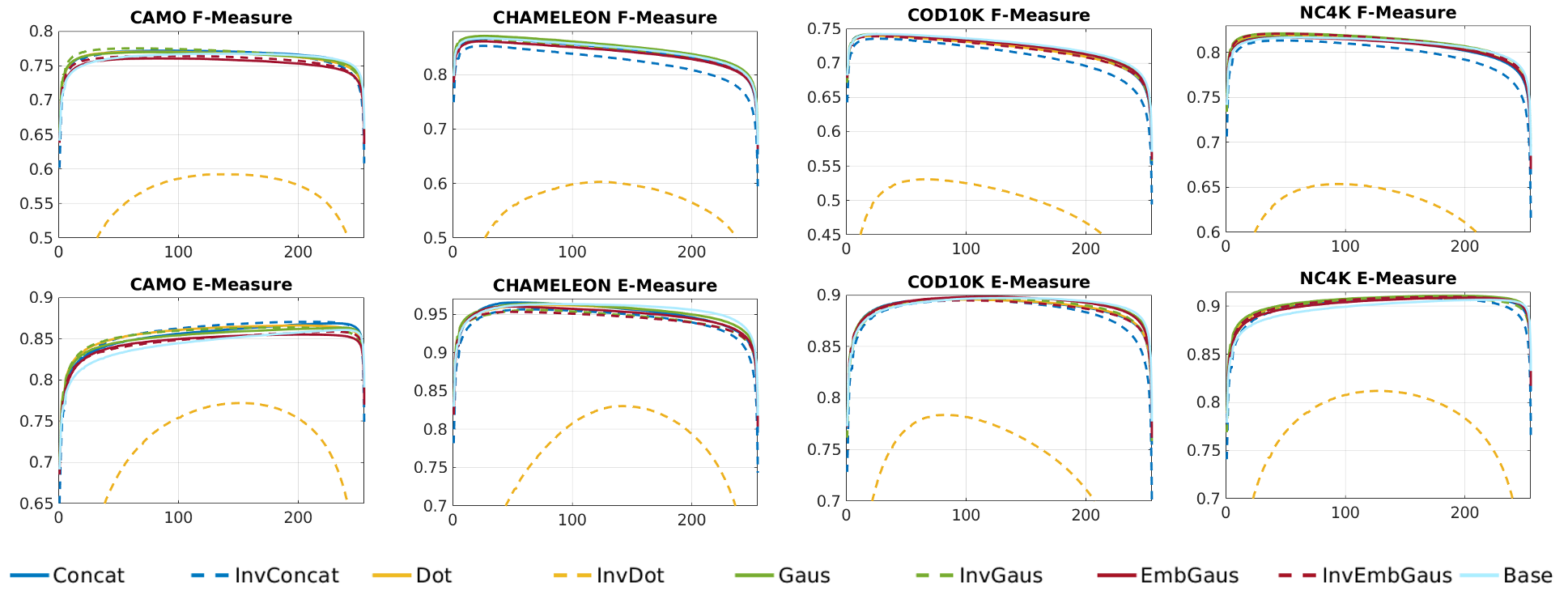} \\
    \end{tabular}
    \caption{{F-measure and E-measure curves on four benchmark camouflage testing datasets. Best view in color.}}
    \label{fig:f_e_measure}
\end{figure}

\begin{table}[h]
  \centering
  \scriptsize
  \renewcommand{\arraystretch}{1.3}
  \renewcommand{\tabcolsep}{1.1mm}
  \caption{Performance comparison of the invertible attention with non-invertible attention for camouflaged object detection. \enquote{Base} denotes the baseline model without attention. \enquote{Concat}, \enquote{Dot}, \enquote{Gaus} and \enquote{EGaus} represent the four attention modules, \ie Concatenation, Dot-product, Gaussian and Embedded Gaussian. \enquote{Inv-} denotes their corresponding invertible attention models. 
}
  \begin{tabular}{l|cccc|cccc|cccc|cccc}
  \hline
  &\multicolumn{4}{c|}{CAMO~\cite{le2019anabranch}}&\multicolumn{4}{c|}{CHAMELEON~\cite{Chameleon2018}}&\multicolumn{4}{c|}{COD10K~\cite{fan2020camouflaged}}&\multicolumn{4}{c}{NC4K~\cite{yunqiu_cod21}} \\
    Method & $S_{\alpha}\uparrow$&$F_{\beta}\uparrow$&$E_{\xi}\uparrow$&$\mathcal{M}\downarrow$& $S_{\alpha}\uparrow$&$F_{\beta}\uparrow$&$E_{\xi}\uparrow$&$\mathcal{M}\downarrow$& $S_{\alpha}\uparrow$&$F_{\beta}\uparrow$&$E_{\xi}\uparrow$&$\mathcal{M}\downarrow$& $S_{\alpha}\uparrow$&$F_{\beta}\uparrow$&$E_{\xi}\uparrow$&$\mathcal{M}\downarrow$  \\
  \hline
  Concat & 0.799 & 0.762 & 0.850 & 0.076 & 0.894 & 0.840 & 0.947 & 0.027 & 0.811 & 0.718 & 0.884 & 0.035 & 0.845 & 0.804 & 0.900 & 0.046 \\
  \rowcolor{mygray}
  InvConcat & 0.799 & 0.759 & 0.853 & 0.075 & 0.883 & 0.821 & 0.938 & 0.031 & 0.807 & 0.707 & 0.879 & 0.037 & 0.842 & 0.796 & 0.896 & 0.048 \\
  Dot & 0.799 & 0.760 & 0.853 & 0.076 & 0.895 & 0.846 & 0.948 & 0.027 & 0.810 & 0.717 & 0.884 & 0.035 & 0.845 & 0.806 & 0.900 & 0.046 \\
  \rowcolor{mygray}
  InvDot & 0.686 & 0.538 & 0.696 & 0.147 & 0.711 & 0.544 & 0.739 & 0.109 & 0.680 & 0.480 & 0.716 & 0.091 & 0.732 & 0.606 & 0.755 & 0.108  \\
  Gaus & 0.798 & 0.761 & 0.849 & 0.076 & 0.899 & 0.847 & 0.949 & 0.026 & 0.812 & 0.721 & 0.887 & 0.034 & 0.846 & 0.809 & 0.901 & 0.046 \\
  \rowcolor{mygray}
  InvGaus & 0.801 & 0.764 & 0.852& 0.076 & 0.894 & 0.840 & 0.941 & 0.027 & 0.812 & 0.720 & 0.884 & 0.036 & 0.844 & 0.805 & 0.898 & 0.047  \\
  EGaus & 0.792 & 0.750 & 0.842 & 0.077 & 0.889 & 0.837 & 0.945 & 0.029 & 0.811 & 0.720 & 0.887 & 0.034 & 0.844 & 0.804 & 0.898 & 0.047 \\
  \rowcolor{mygray}
  InvEGaus & 0.793 & 0.754 & 0.843 & 0.078 & 0.887 & 0.838 & 0.939 & 0.031 & 0.809 & 0.717 & 0.883 & 0.036 & 0.845 & 0.808 & 0.900 & 0.047  \\
  \hline
  Base & 0.791 & 0.758 & 0.840 & 0.078 & 0.894 & 0.843 & 0.951 & 0.028 & 0.811 & 0.723 & 0.887 & 0.035 & 0.841 & 0.806 & 0.894 & 0.048 \\
   \hline
  \end{tabular}
  \label{tab:invertible-attention-performance}
\end{table}

\subsection{Invertible Attention \emph{vs.} Non-invertible Attention in Discriminative Learning}

\textbf{Implementation:} We compare the performance of the invertible attention with non-invertible attention in a dense prediction task: camouflaged object detection \cite{skelhorn2016cognition,merilaita2017camouflage, background_matching}, which aims to accurately localize the whole scope of the camouflaged objects. Specifically, we adopt the camouflaged object detection network from \cite{yunqiu_cod21}, which takes ResNet50 \cite{he2015deep} as the backbone. The model is trained to generate a one channel camouflage map, representing the possibility of each pixel belong to a camouflaged object. We train the model on the training set of the COD10K dataset \cite{fan2020camouflaged}, and evaluate on four public camouflage testing datasets, including CAMO \cite{le2019anabranch}, CHAMELEON \cite{Chameleon2018}, COD10K testing dataset and NC4K testing dataset \cite{yunqiu_cod21}. We use four evaluation metrics to evaluate model performance, including Mean Absolute Error, Mean F-measure, Mean E-measure \cite{fan2018enhanced} and S-measure \cite{fan2017structure} denoted as $\mathcal{M}$, $F_\beta$, $E_\xi$, $S_{\alpha}$, respectively.
Details about the networks, datasets and evaluation metrics are provided in Appendix \ref{sec: supp discriminative}.

\noindent\textbf{Quantitative results:} To test the effectiveness of both our non-invertible and invertible attention models, we add each attention to each level of the backbone network in \cite{yunqiu_cod21}.
We summarize performance of invertible attention and non-invertible attention in Table \ref{tab:invertible-attention-performance}. Given the trade-off between expressive power and invertibility described in Section \ref{sec: practical implementation}, we expect the performance of invertible attention to be lower than their corresponding non-invertible forms, but still higher than baseline model which does not have attention. Actually, we observe slightly decayed performance over different kinds of invertible attention compared to their non-invertible versions.
Besides, we show the F-measure and E-measure curves of these models with different attention models in Fig.~\ref{fig:f_e_measure} on the four benchmarks. We observe that the curves produced by Concatenation, Gaussian and Embedded Gaussian attention are very close to their non-invertible counterparts. The results clearly show the effectiveness of the invertible attention models.

\begin{figure}[h]
 \vspace{2mm}
   \begin{center}
   \begin{tabular}{c@{ }c@{ }c@{ }c@{ }c@{ }c@{ }c@{ }c@{ }c@{ }c@{ }}
   {\includegraphics[width=0.093\linewidth]{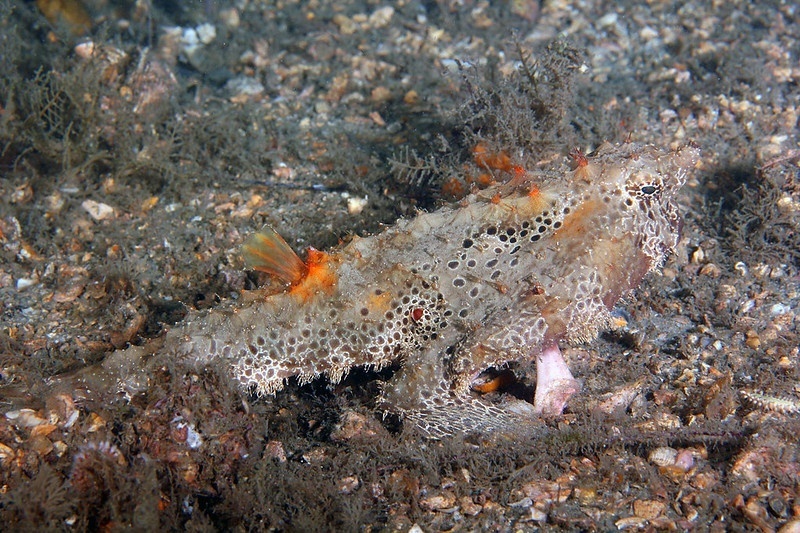}}& 
   {\includegraphics[width=0.093\linewidth]{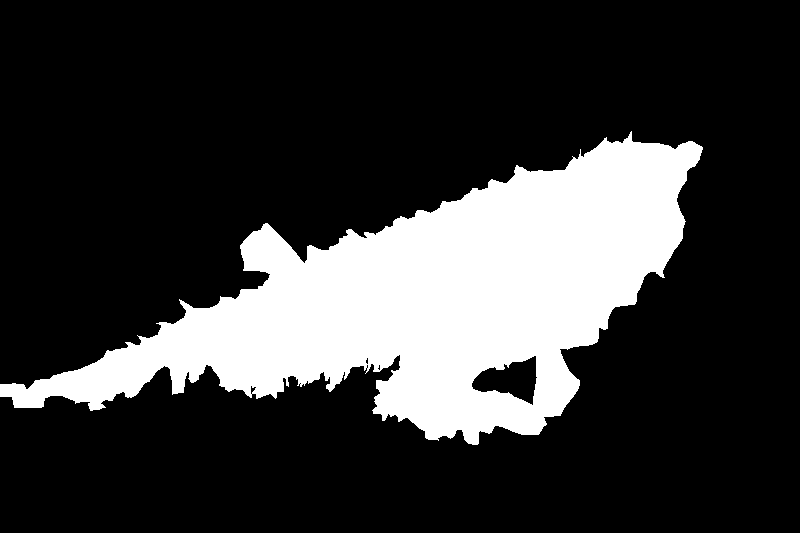}}
   & 
   
   {\includegraphics[width=0.093\linewidth]{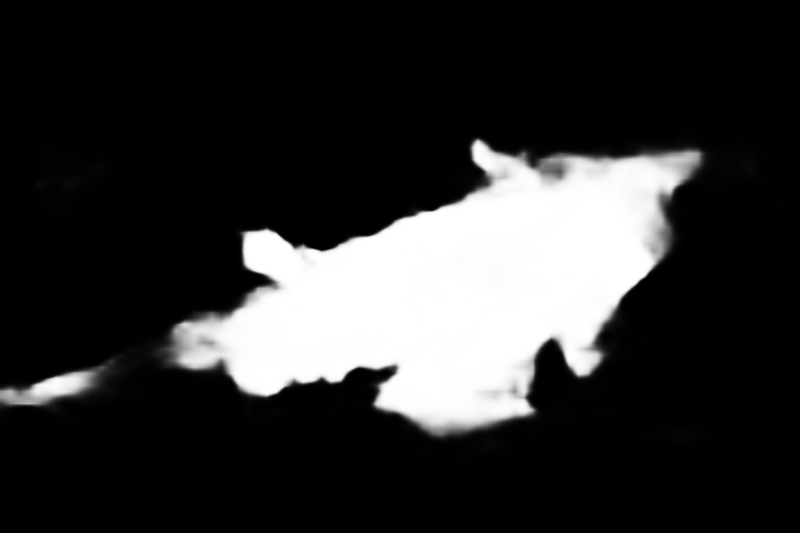}}
   & 
   {\includegraphics[width=0.093\linewidth]{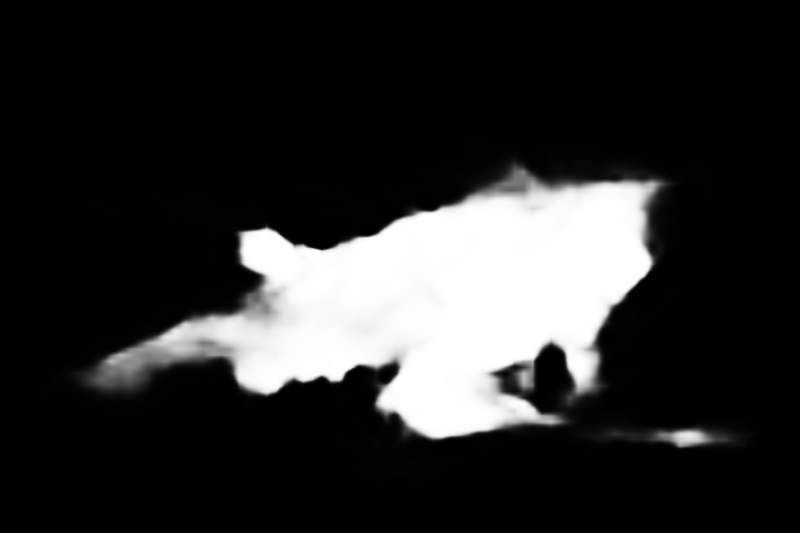}}
   & 
   {\includegraphics[width=0.093\linewidth]{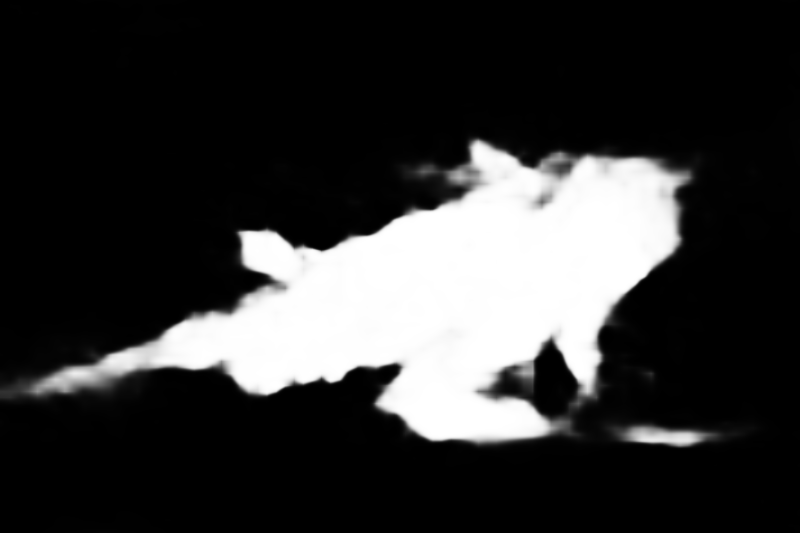}}
   & 
   {\includegraphics[width=0.093\linewidth]{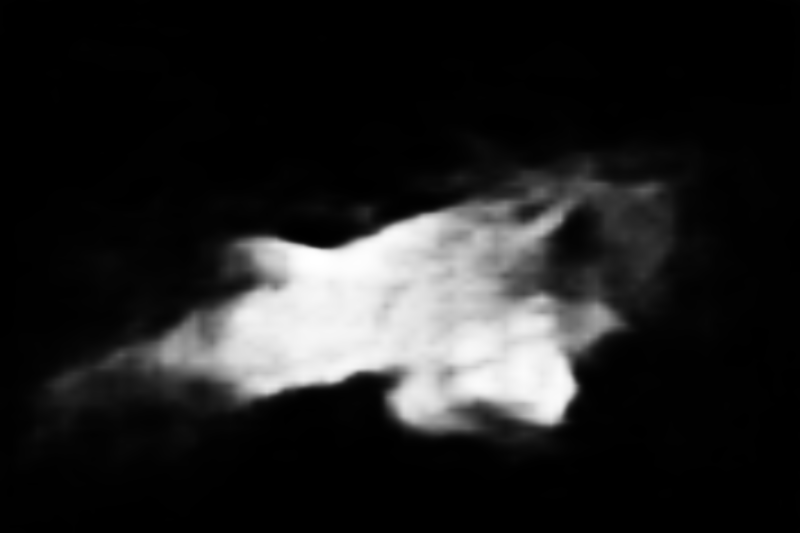}} &
   {\includegraphics[width=0.093\linewidth]{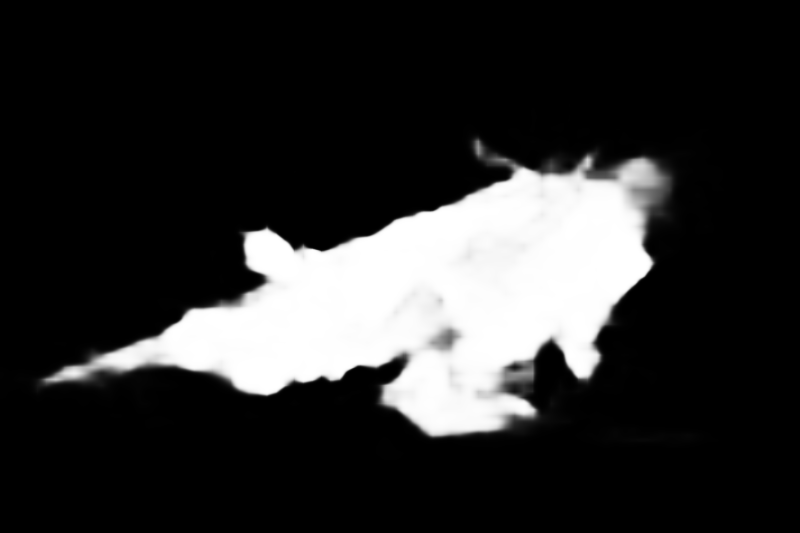}}
   & 
   {\includegraphics[width=0.093\linewidth]{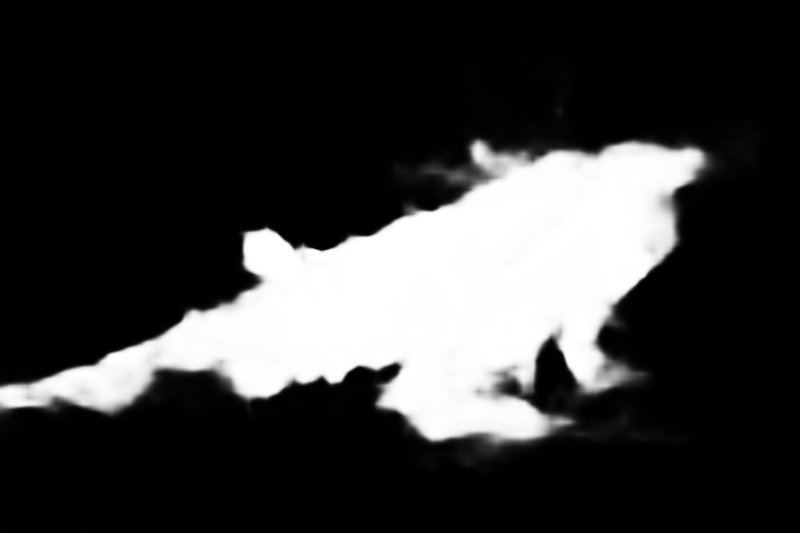}}
   & 
   {\includegraphics[width=0.093\linewidth]{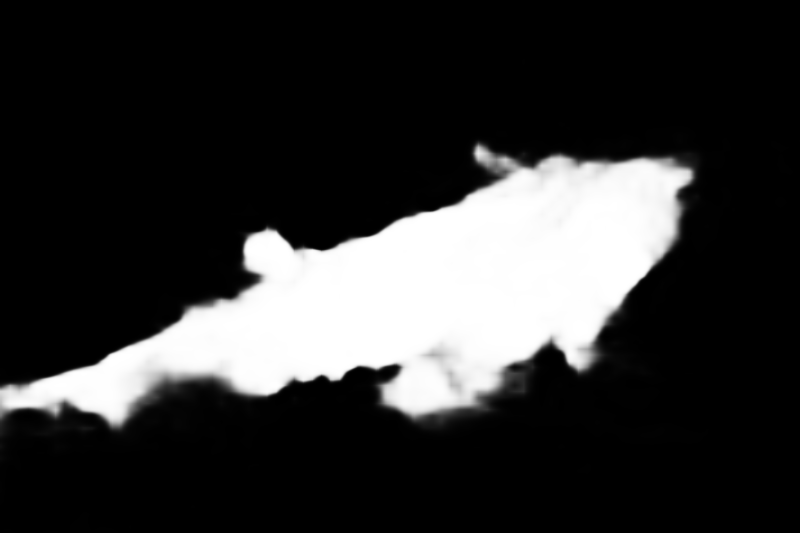}}
   & 
   {\includegraphics[width=0.093\linewidth]{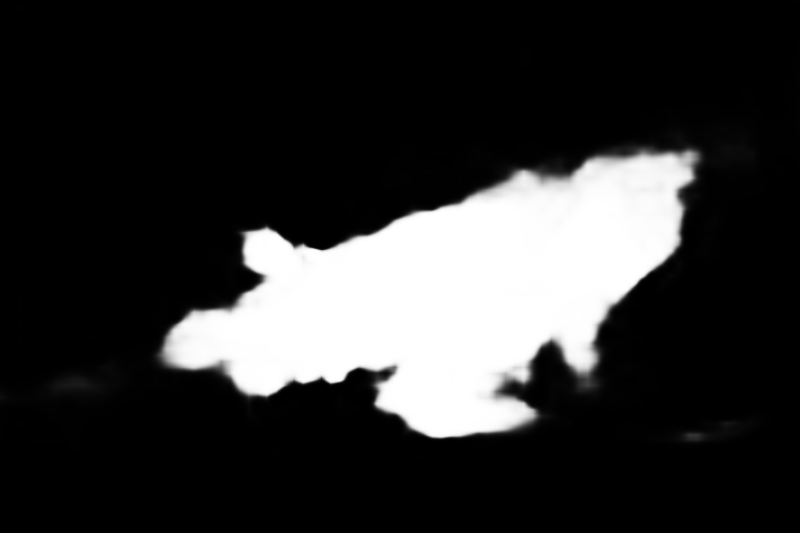}}\\
   {\includegraphics[width=0.093\linewidth]{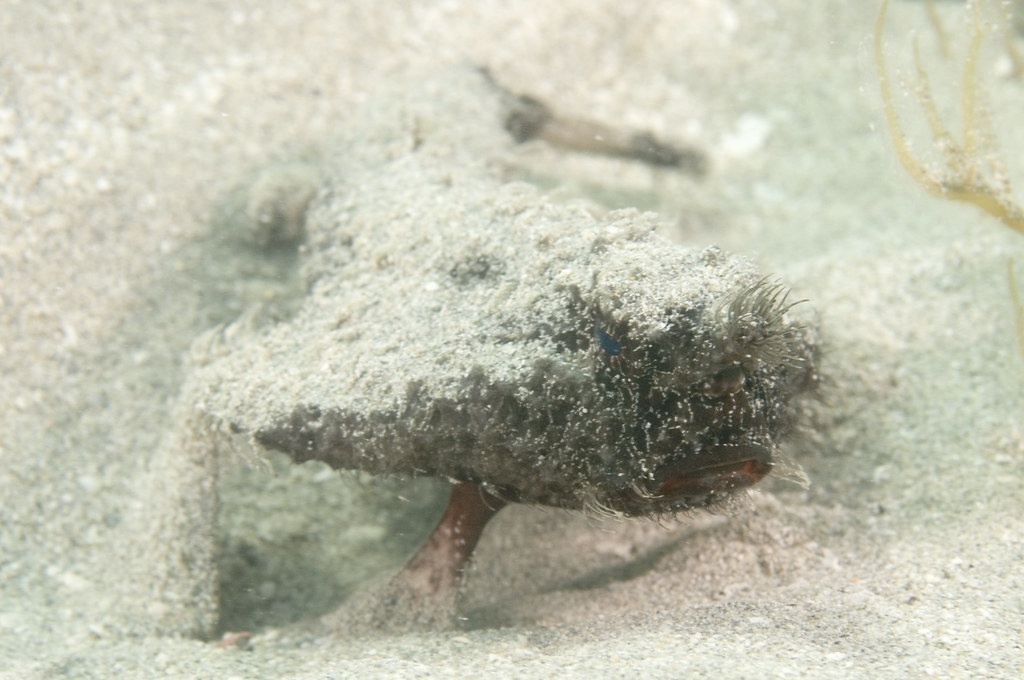}}& 
   {\includegraphics[width=0.093\linewidth]{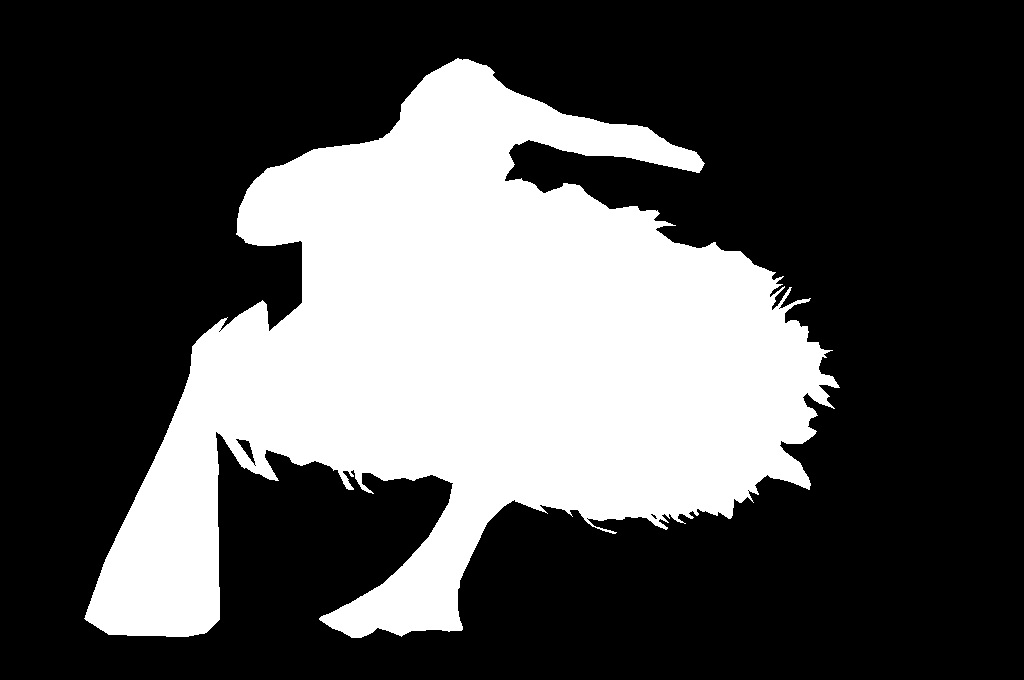}}
   & 
   
   {\includegraphics[width=0.093\linewidth]{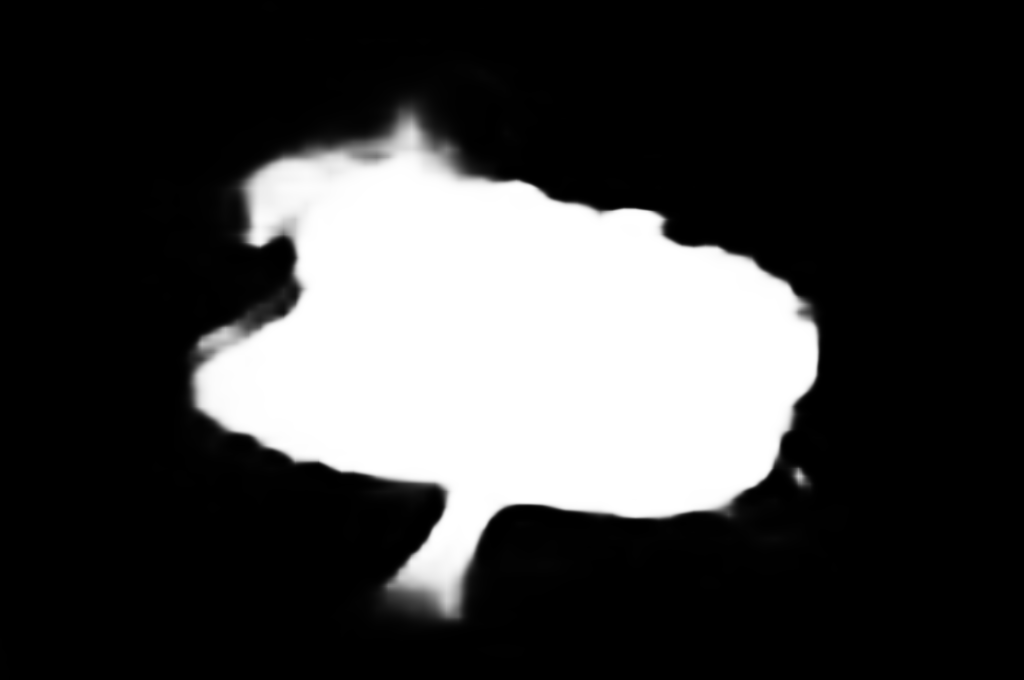}}
   & 
   {\includegraphics[width=0.093\linewidth]{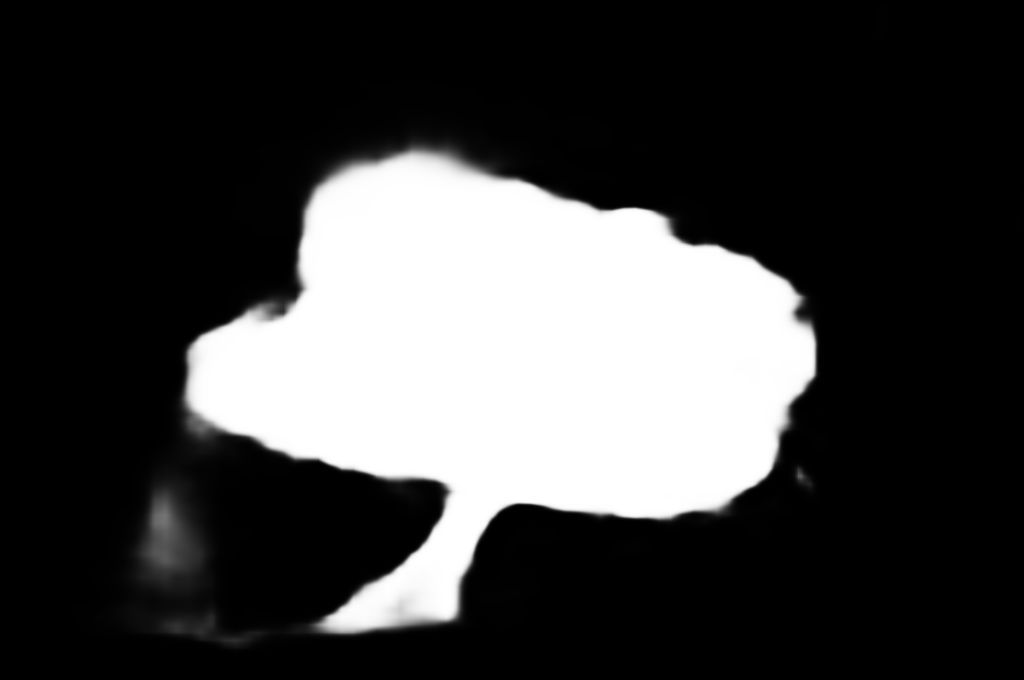}}
   & 
   {\includegraphics[width=0.093\linewidth]{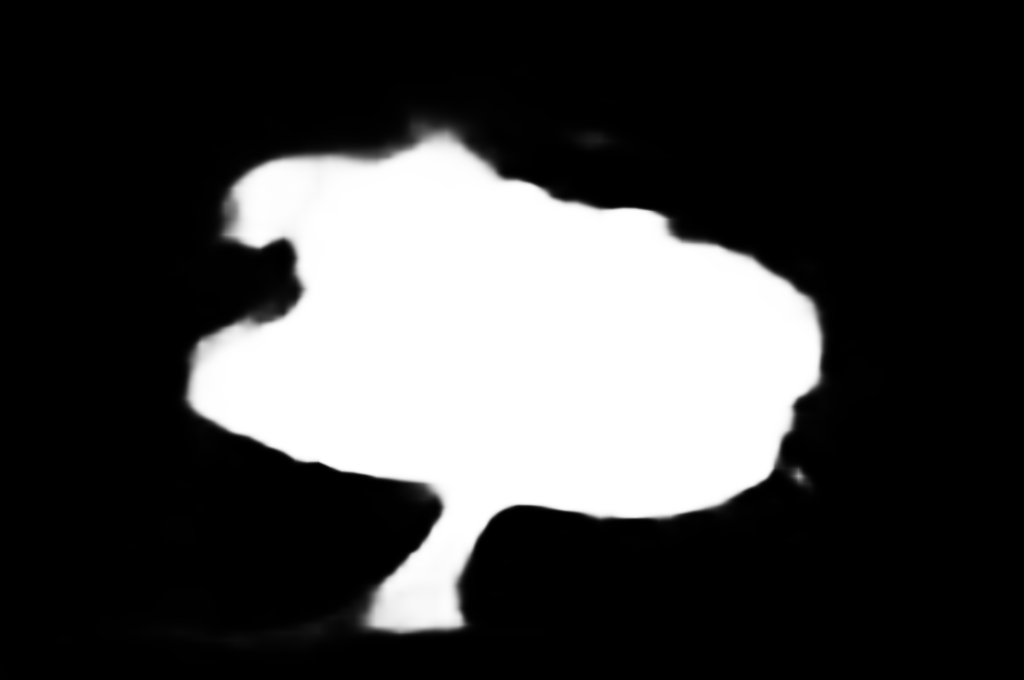}}
   & 
   {\includegraphics[width=0.093\linewidth]{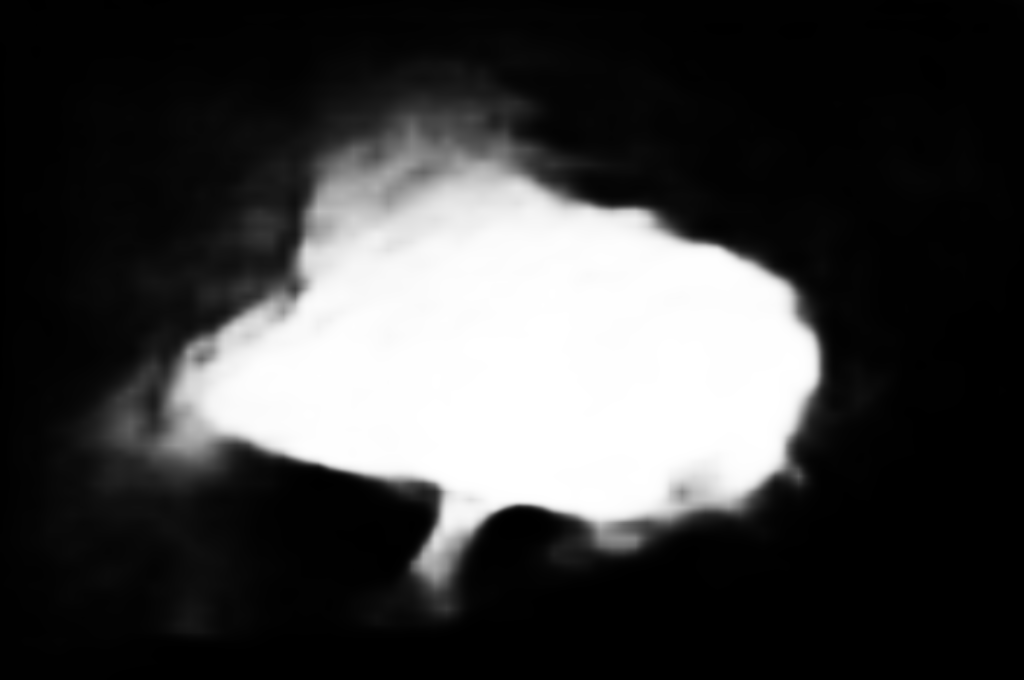}} &
   {\includegraphics[width=0.093\linewidth]{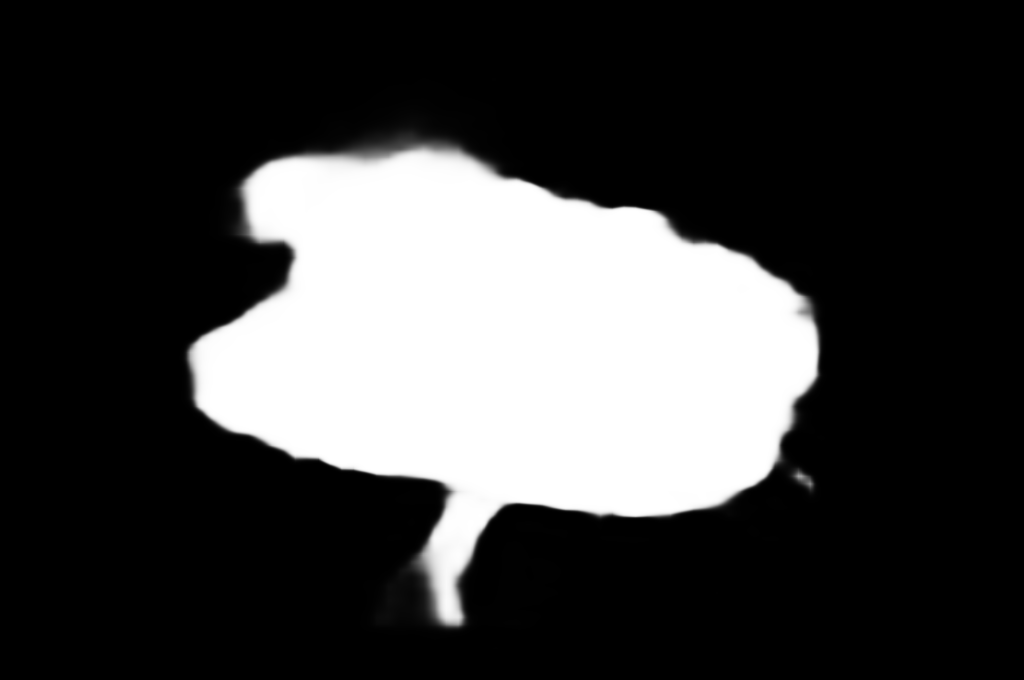}}
   & 
   {\includegraphics[width=0.093\linewidth]{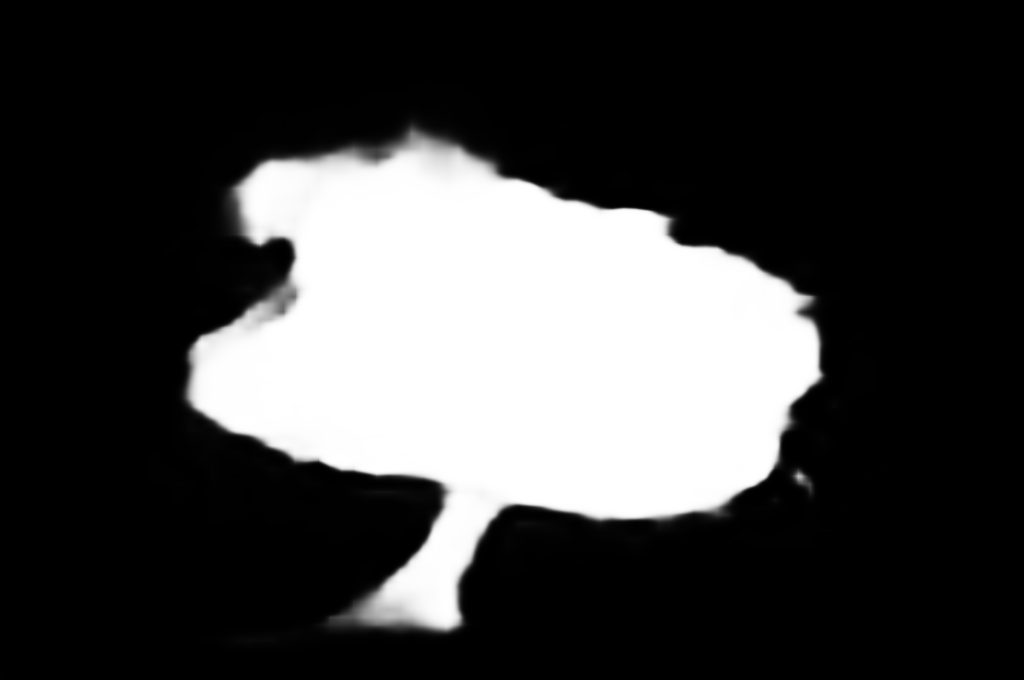}}
   & 
   {\includegraphics[width=0.093\linewidth]{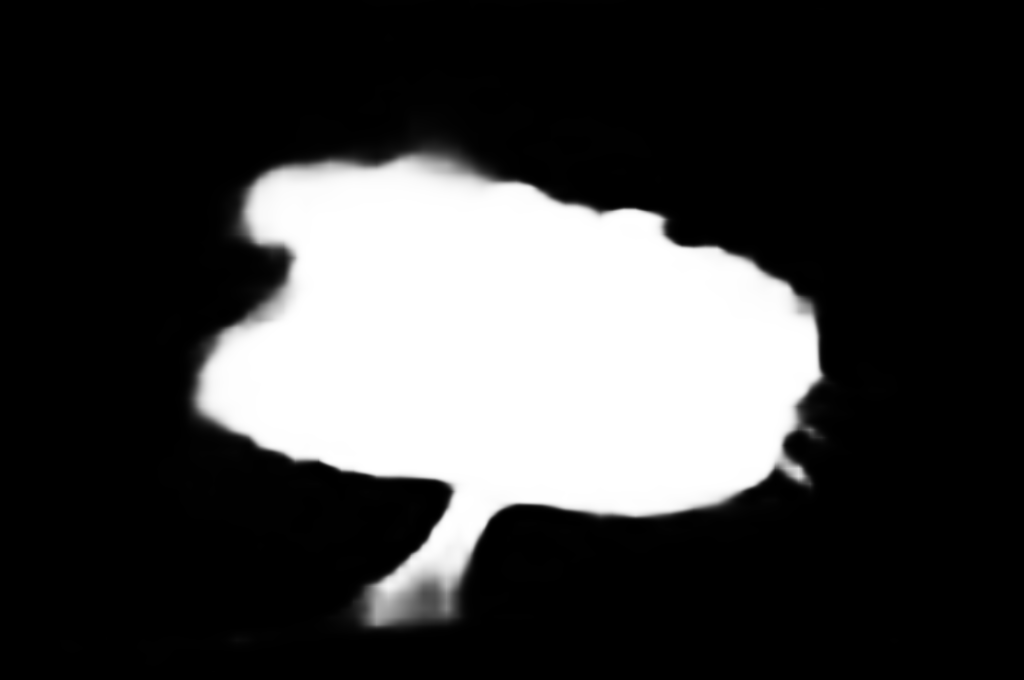}}
   & 
   {\includegraphics[width=0.093\linewidth]{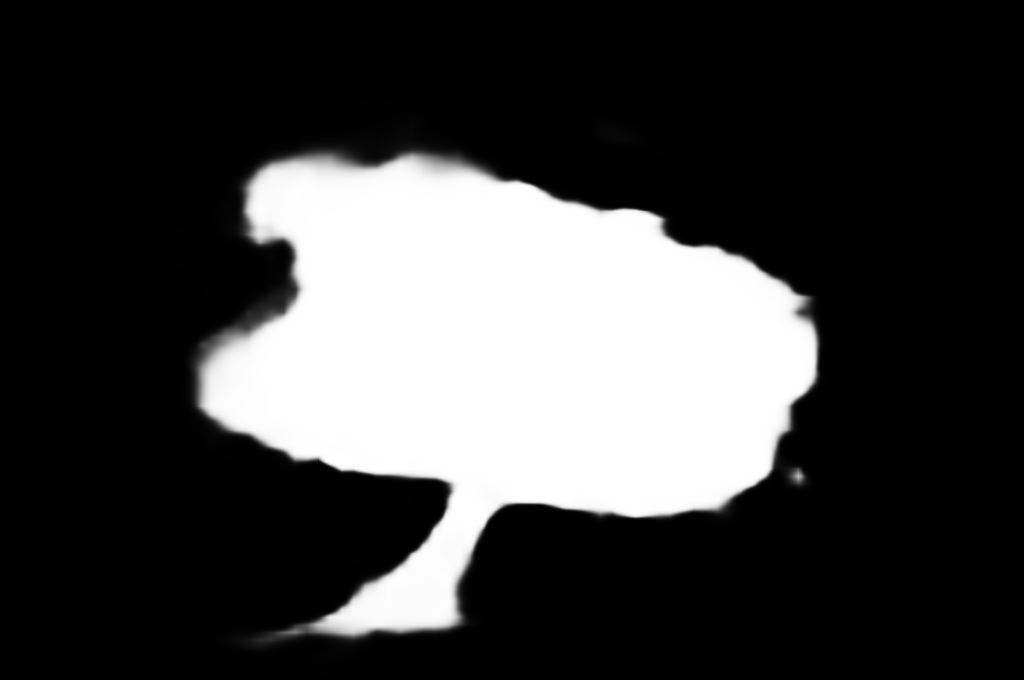}}\\
   {\includegraphics[width=0.093\linewidth]{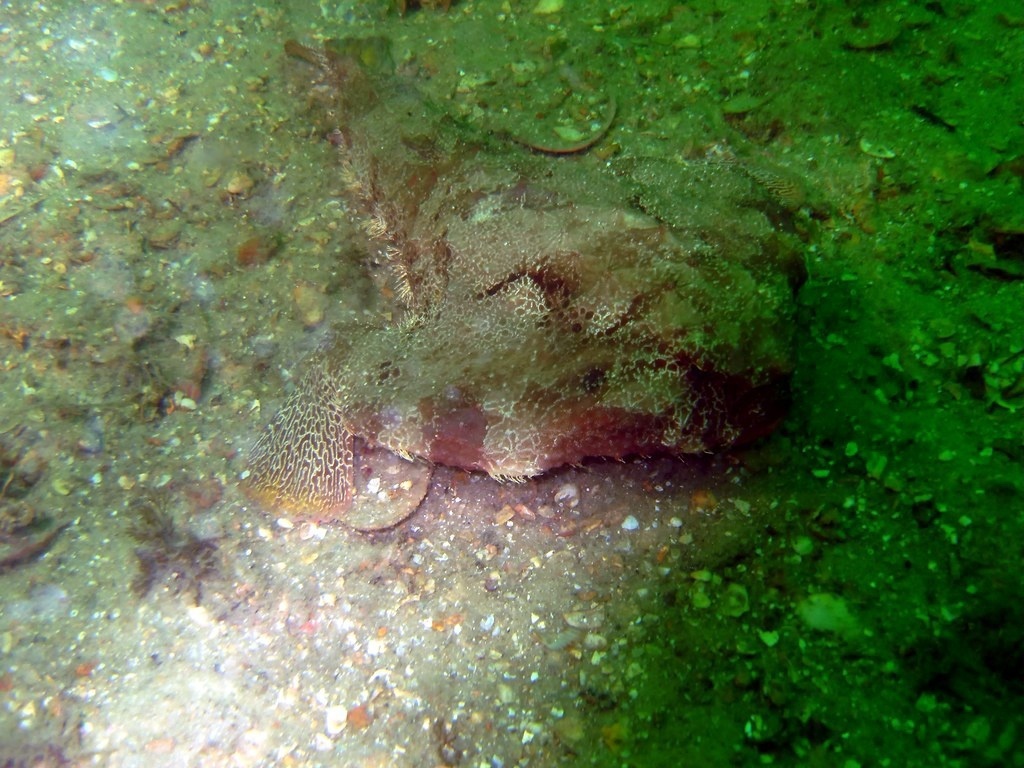}}& 
   {\includegraphics[width=0.093\linewidth]{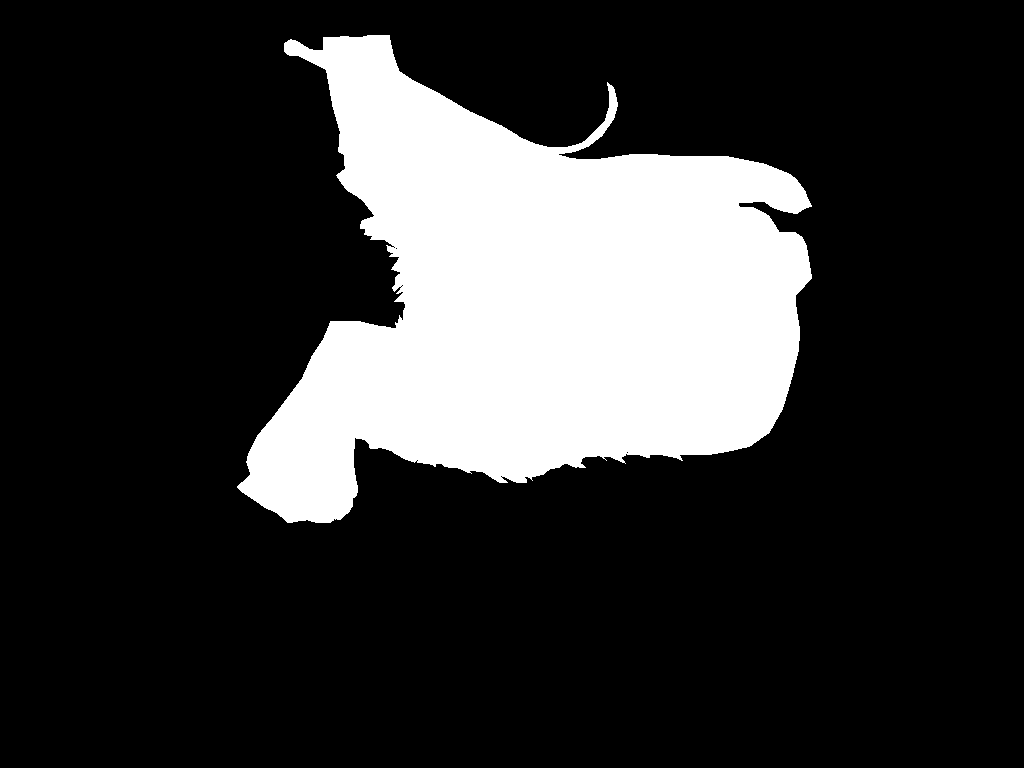}}
   & 
    
   {\includegraphics[width=0.093\linewidth]{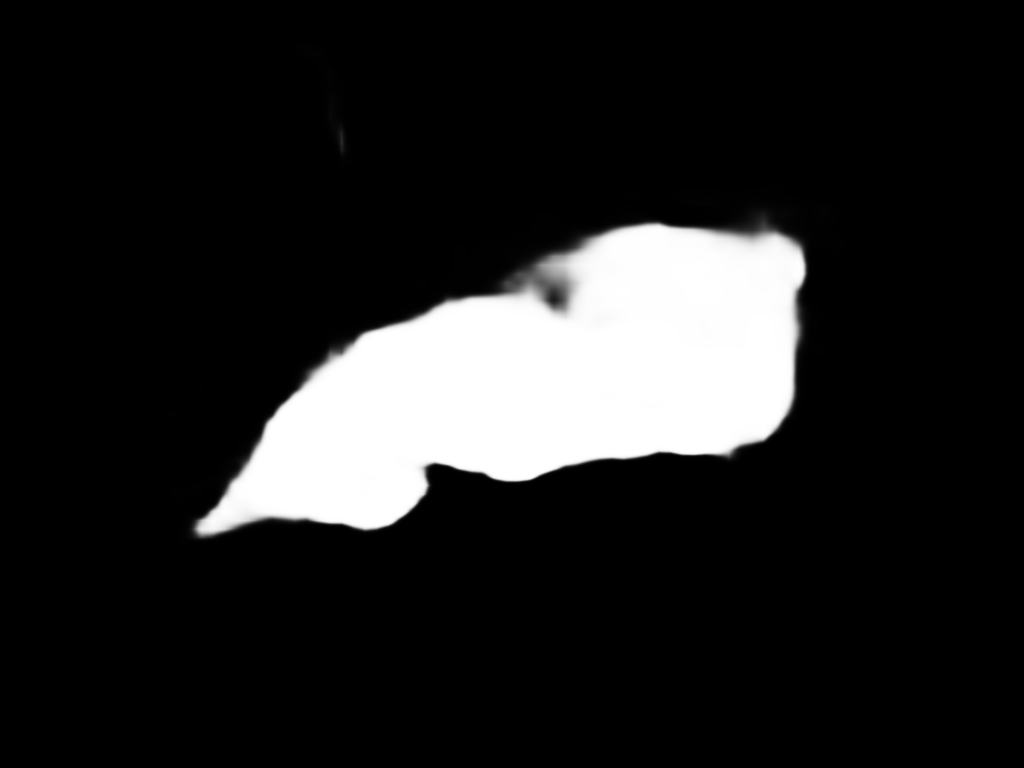}}
   & 
   {\includegraphics[width=0.093\linewidth]{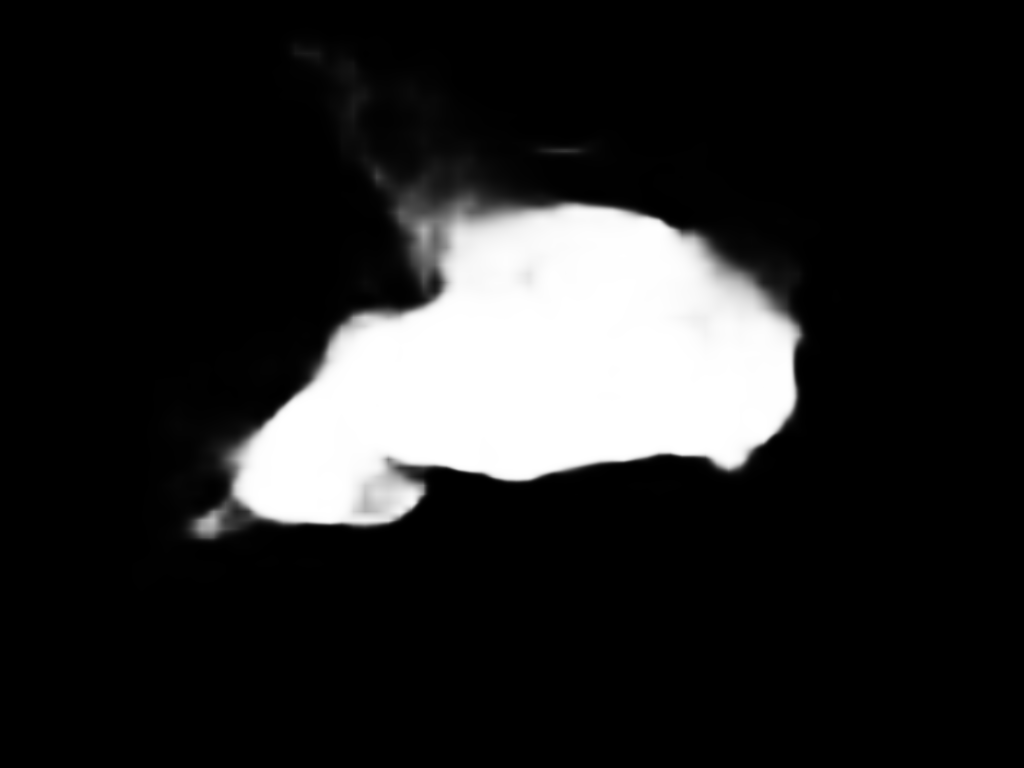}}
   & 
   {\includegraphics[width=0.093\linewidth]{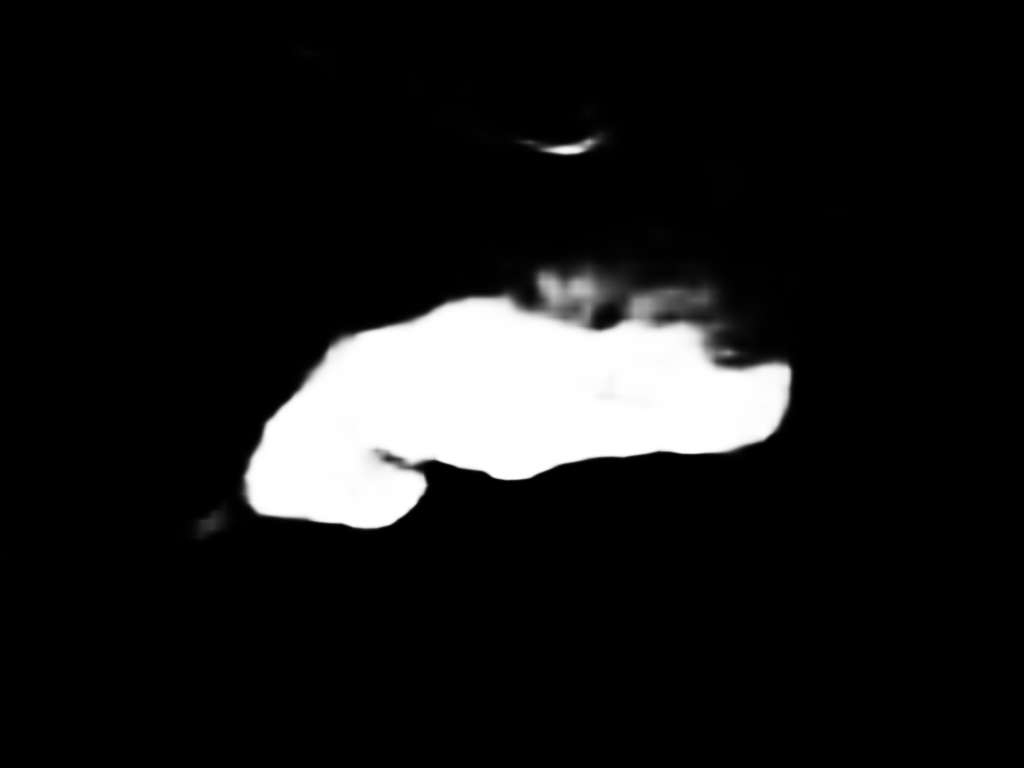}}
   & 
   {\includegraphics[width=0.093\linewidth]{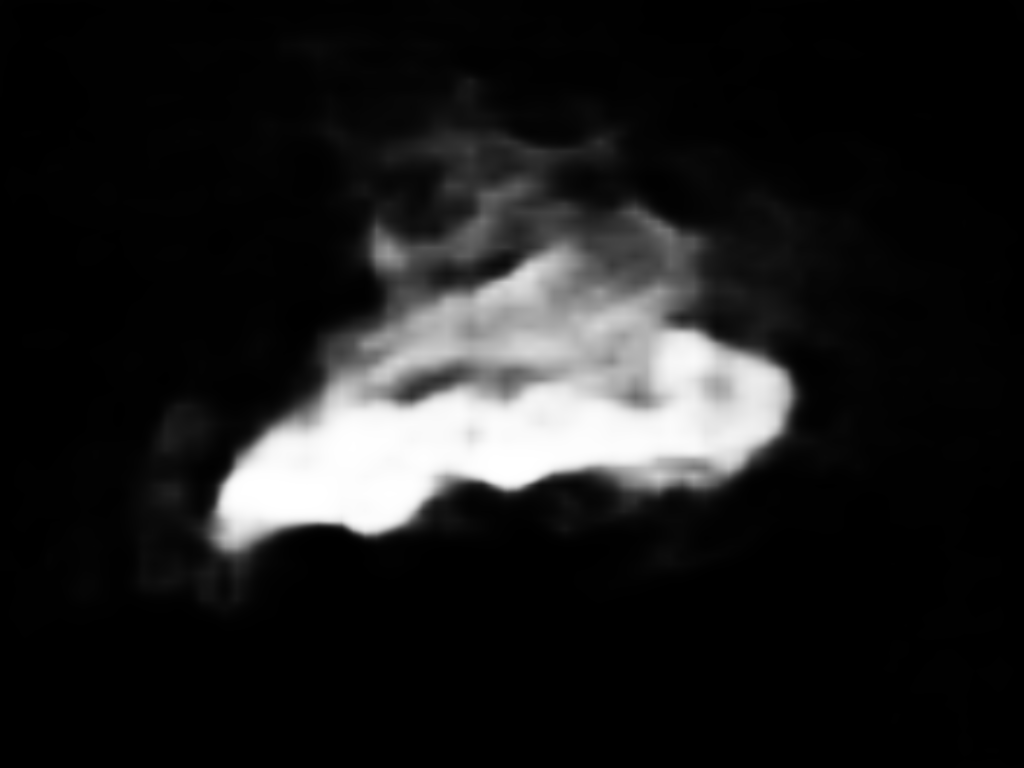}} &
   {\includegraphics[width=0.093\linewidth]{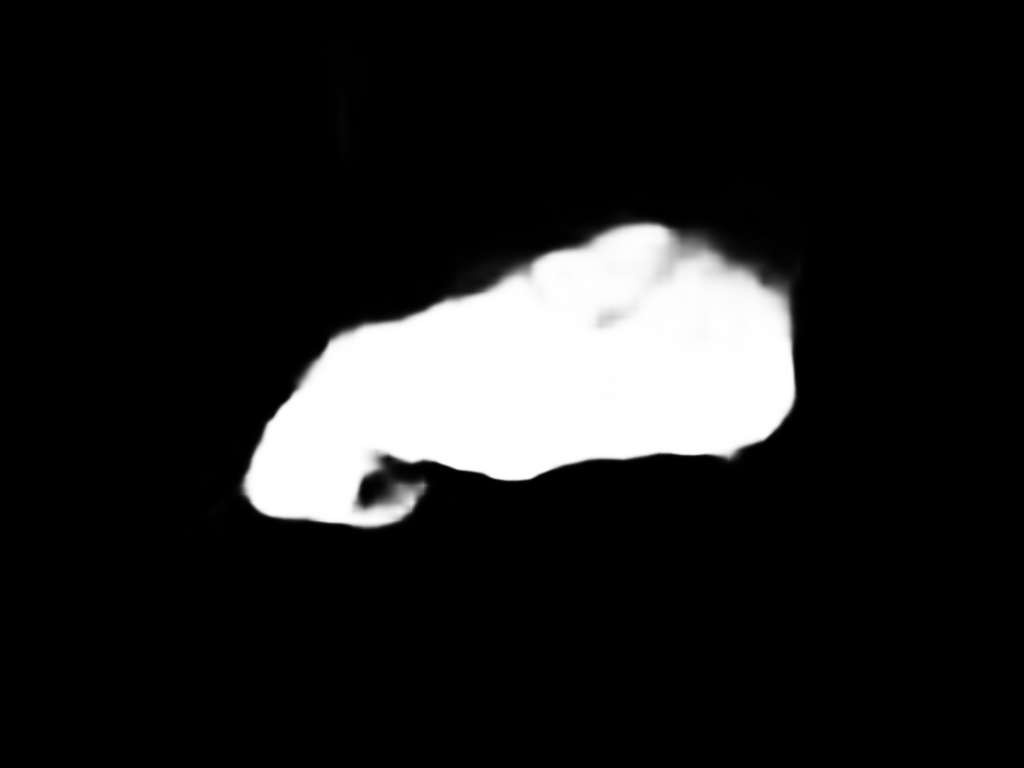}}
   & 
   {\includegraphics[width=0.093\linewidth]{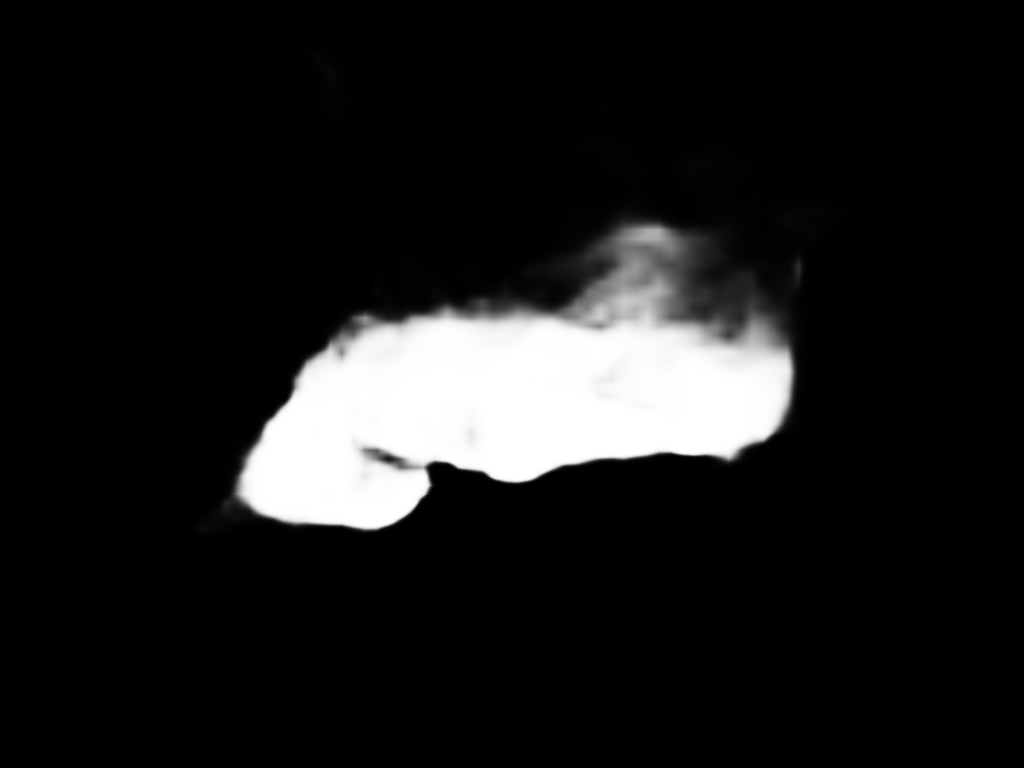}}
   & 
   {\includegraphics[width=0.093\linewidth]{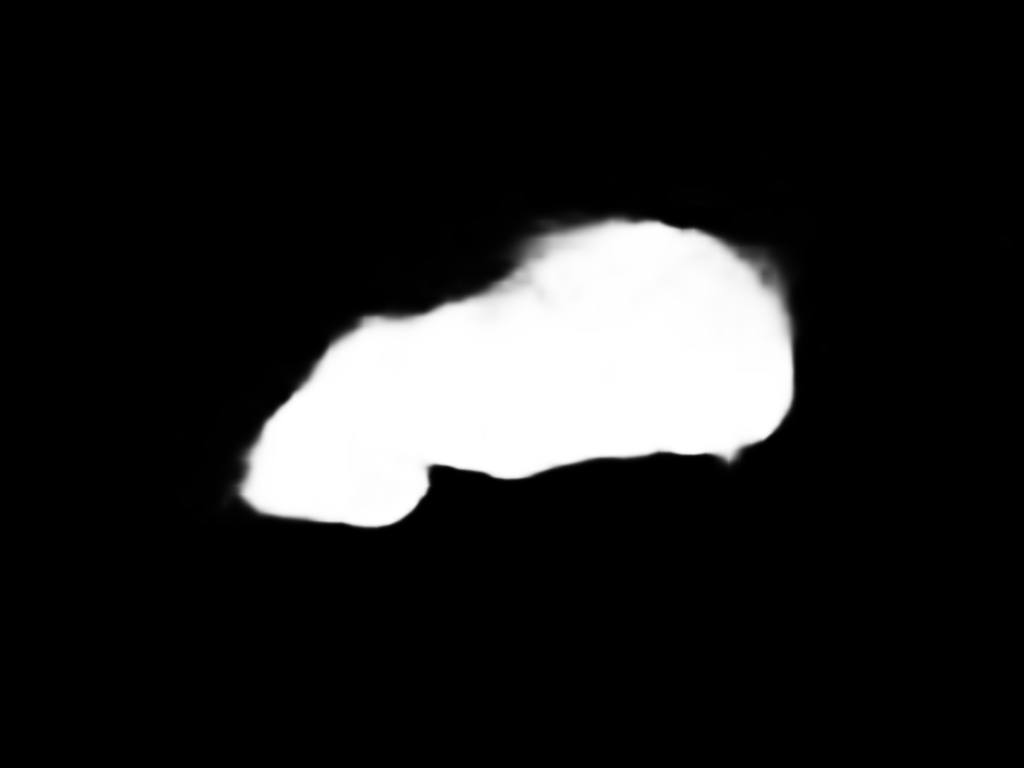}}
   & 
   {\includegraphics[width=0.093\linewidth]{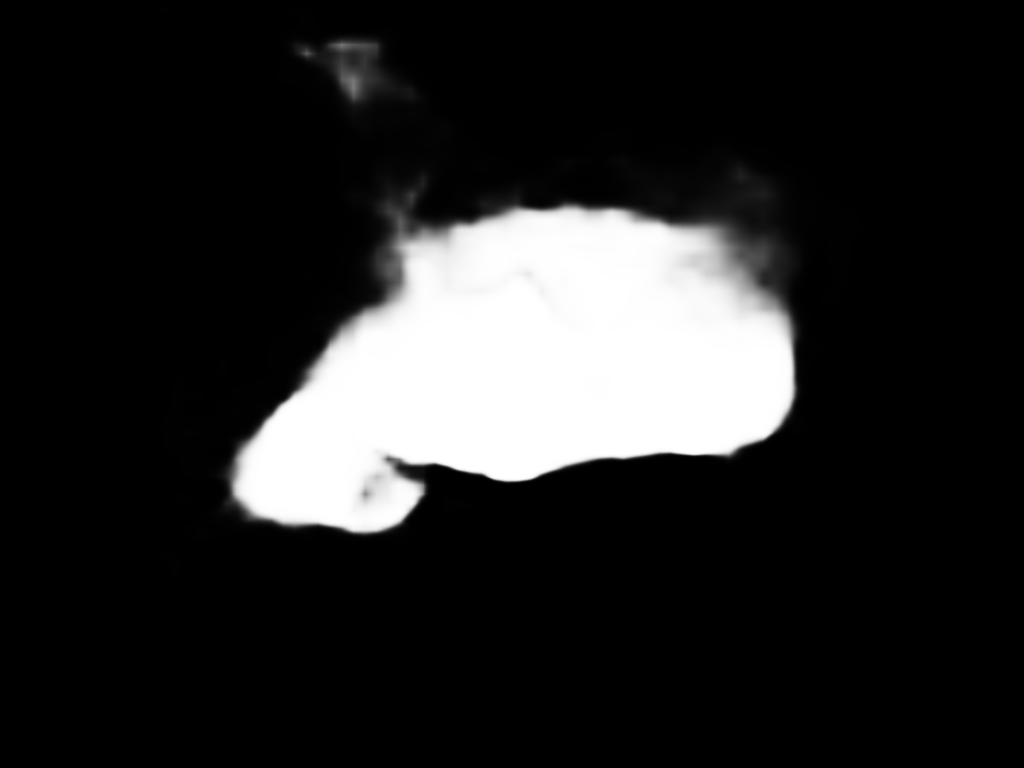}}\\
  {\includegraphics[width=0.093\linewidth]{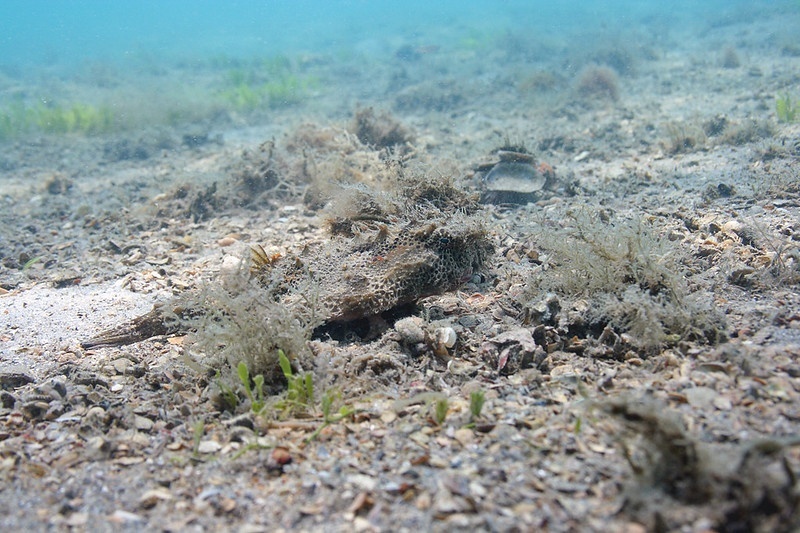}}& 
  {\includegraphics[width=0.093\linewidth]{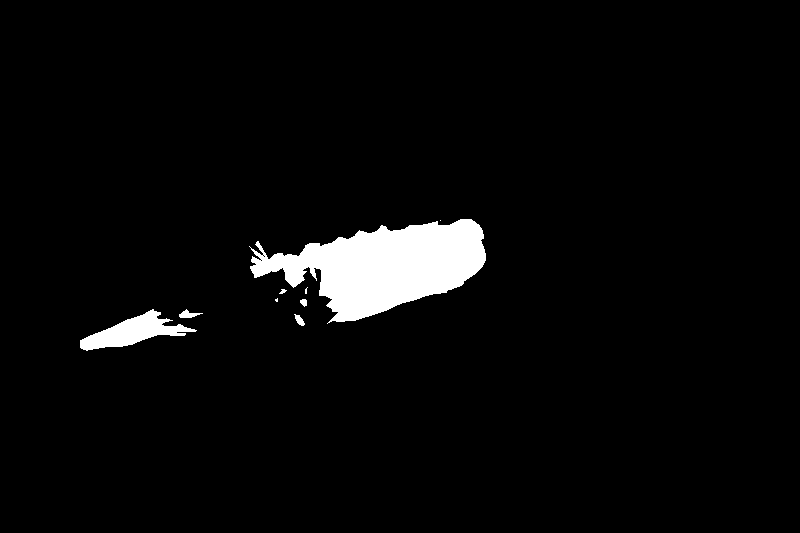}}
  & 
   
  {\includegraphics[width=0.093\linewidth]{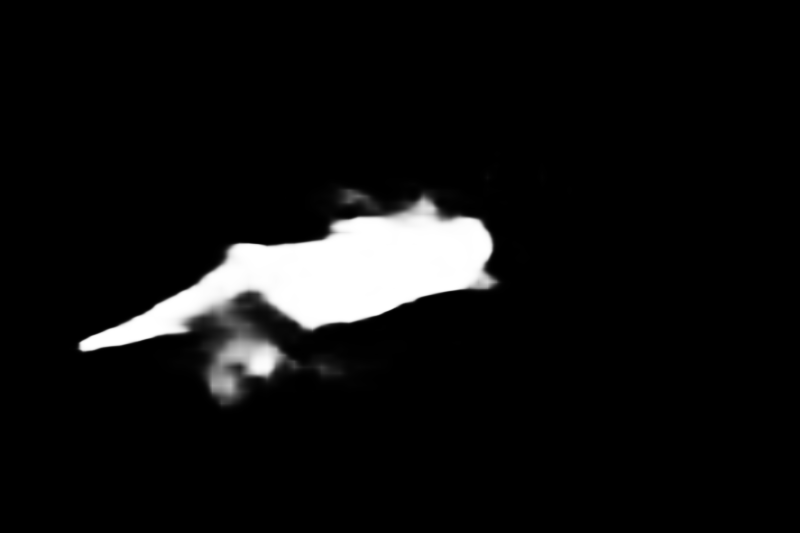}}
  & 
  {\includegraphics[width=0.093\linewidth]{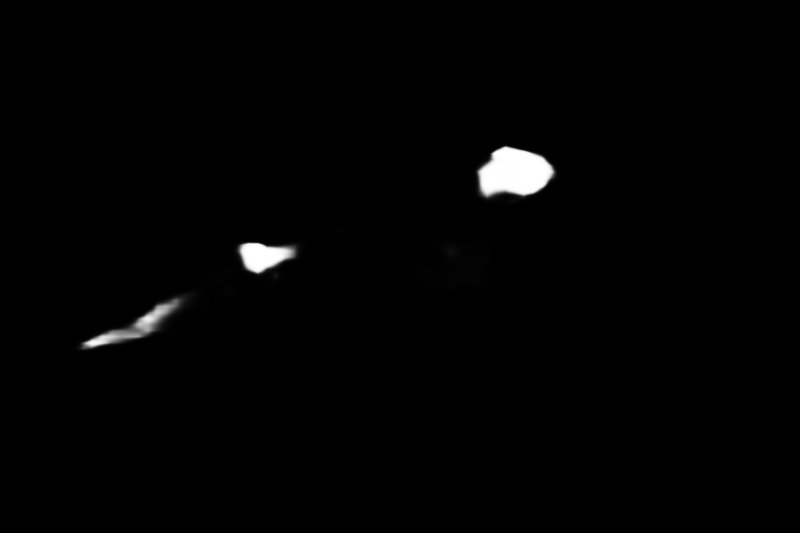}}
  & 
  {\includegraphics[width=0.093\linewidth]{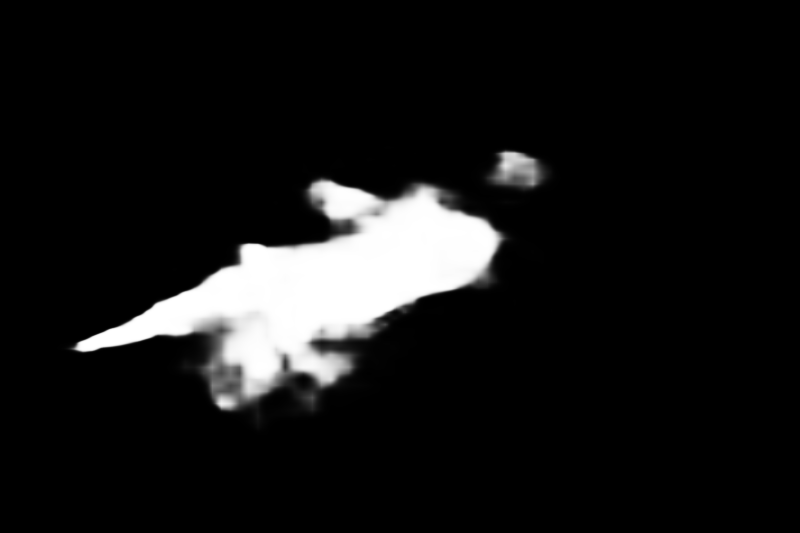}}
  & 
  {\includegraphics[width=0.093\linewidth]{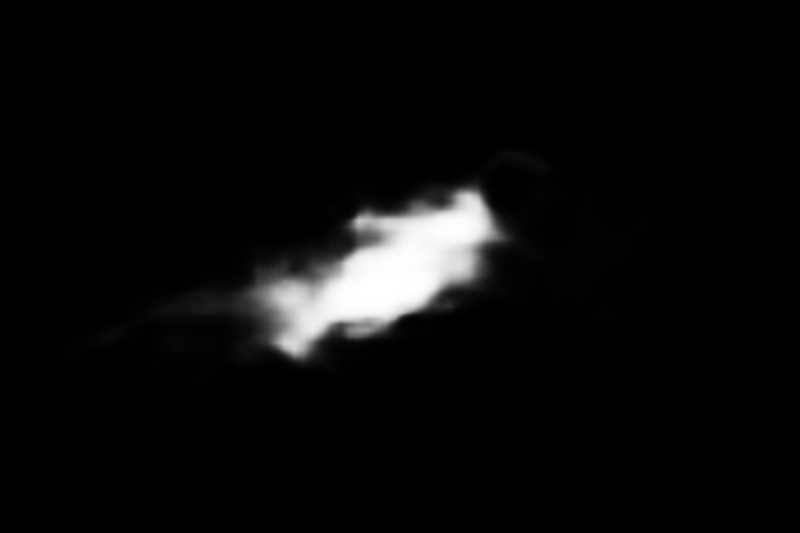}} &
  {\includegraphics[width=0.093\linewidth]{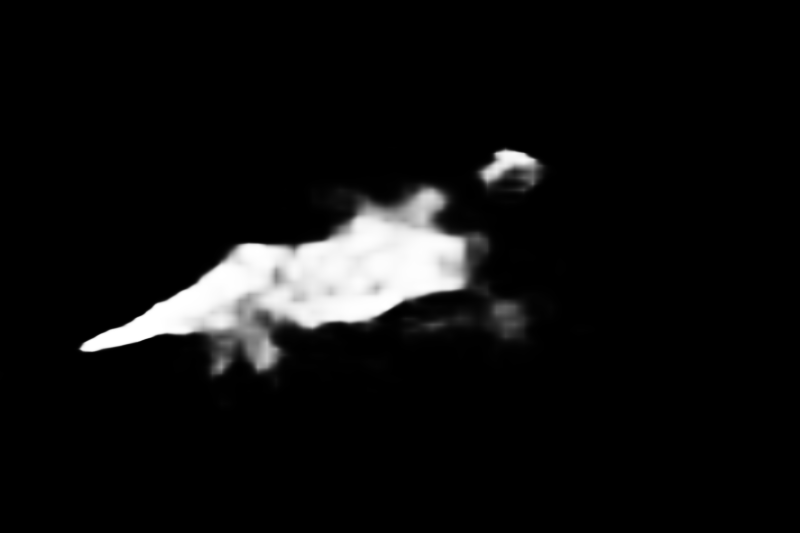}}
  & 
  {\includegraphics[width=0.093\linewidth]{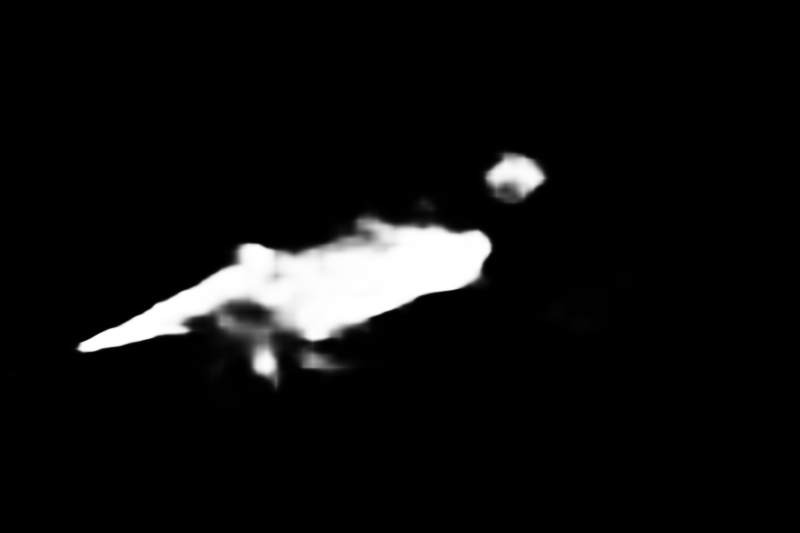}}
  & 
  {\includegraphics[width=0.093\linewidth]{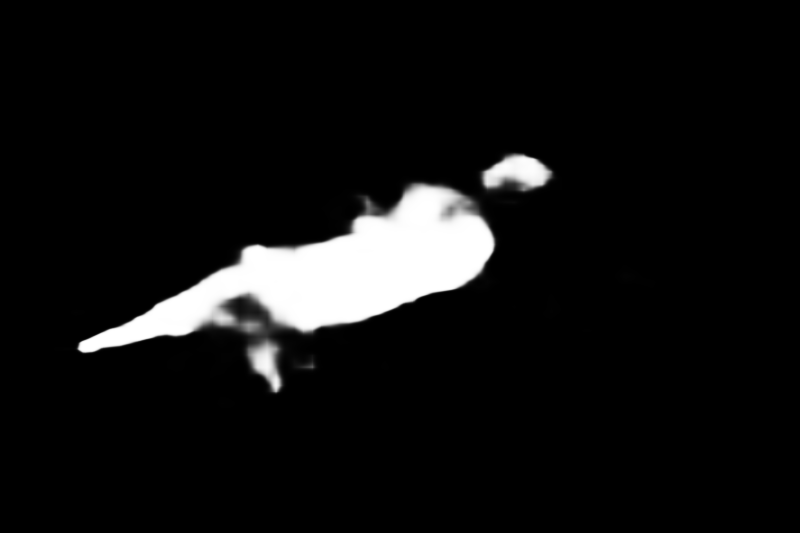}}
  & 
  {\includegraphics[width=0.093\linewidth]{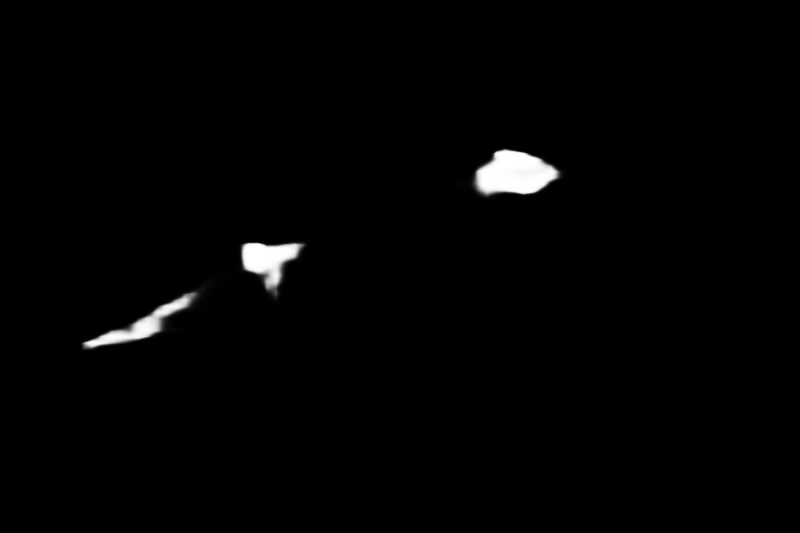}}\\
  {\includegraphics[width=0.093\linewidth]{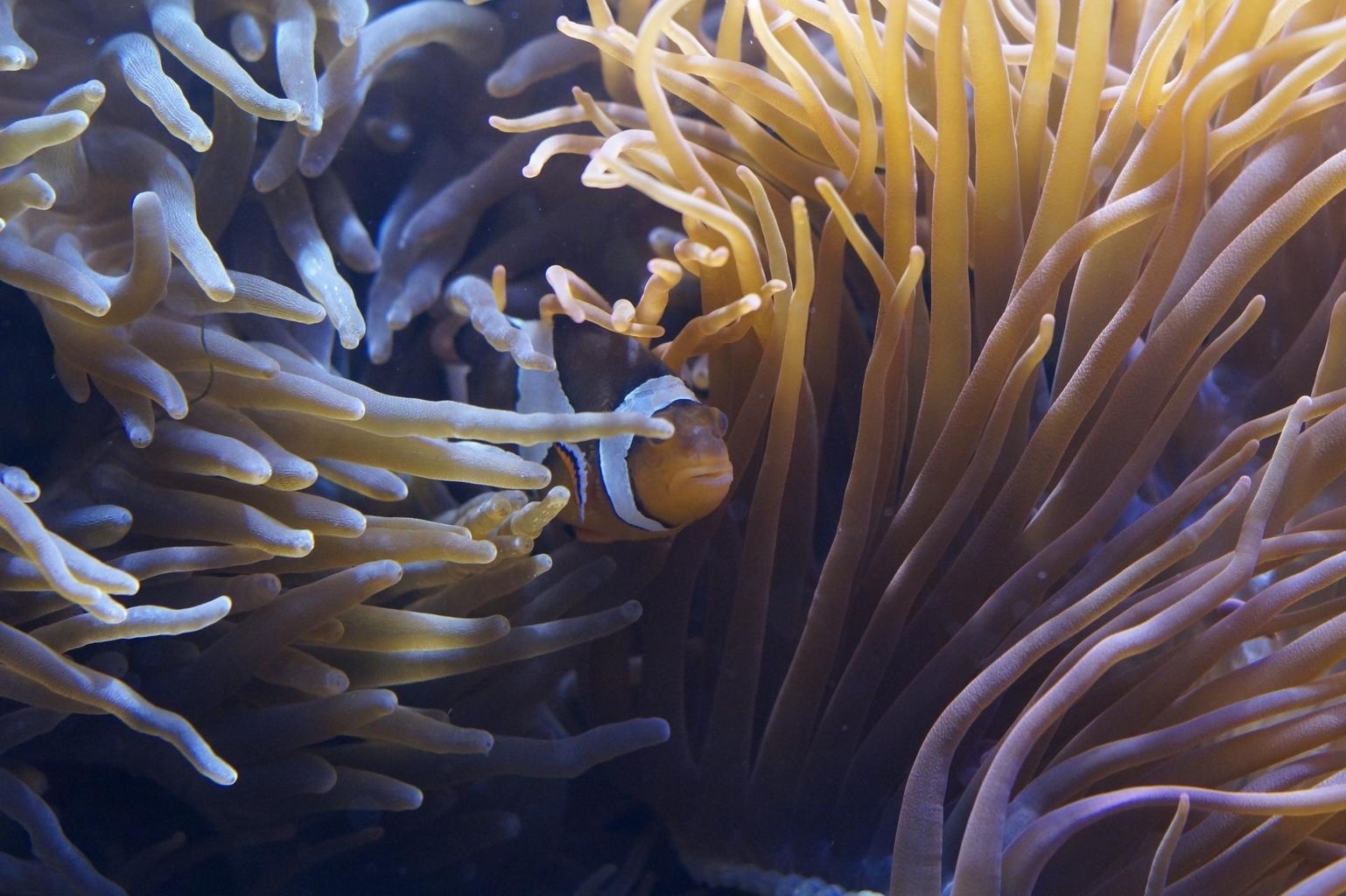}}& 
  {\includegraphics[width=0.093\linewidth]{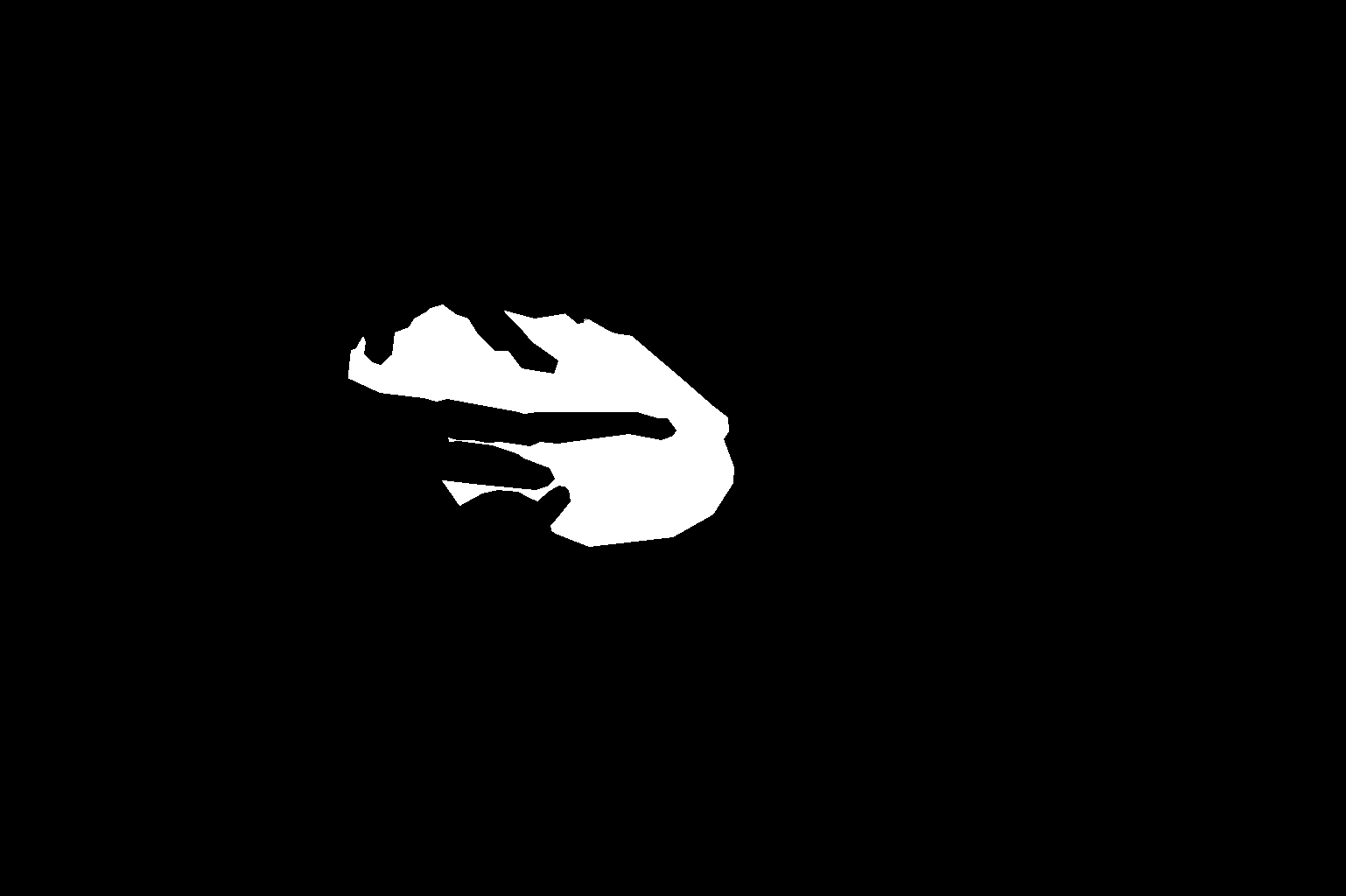}}
  & 
   
  {\includegraphics[width=0.093\linewidth]{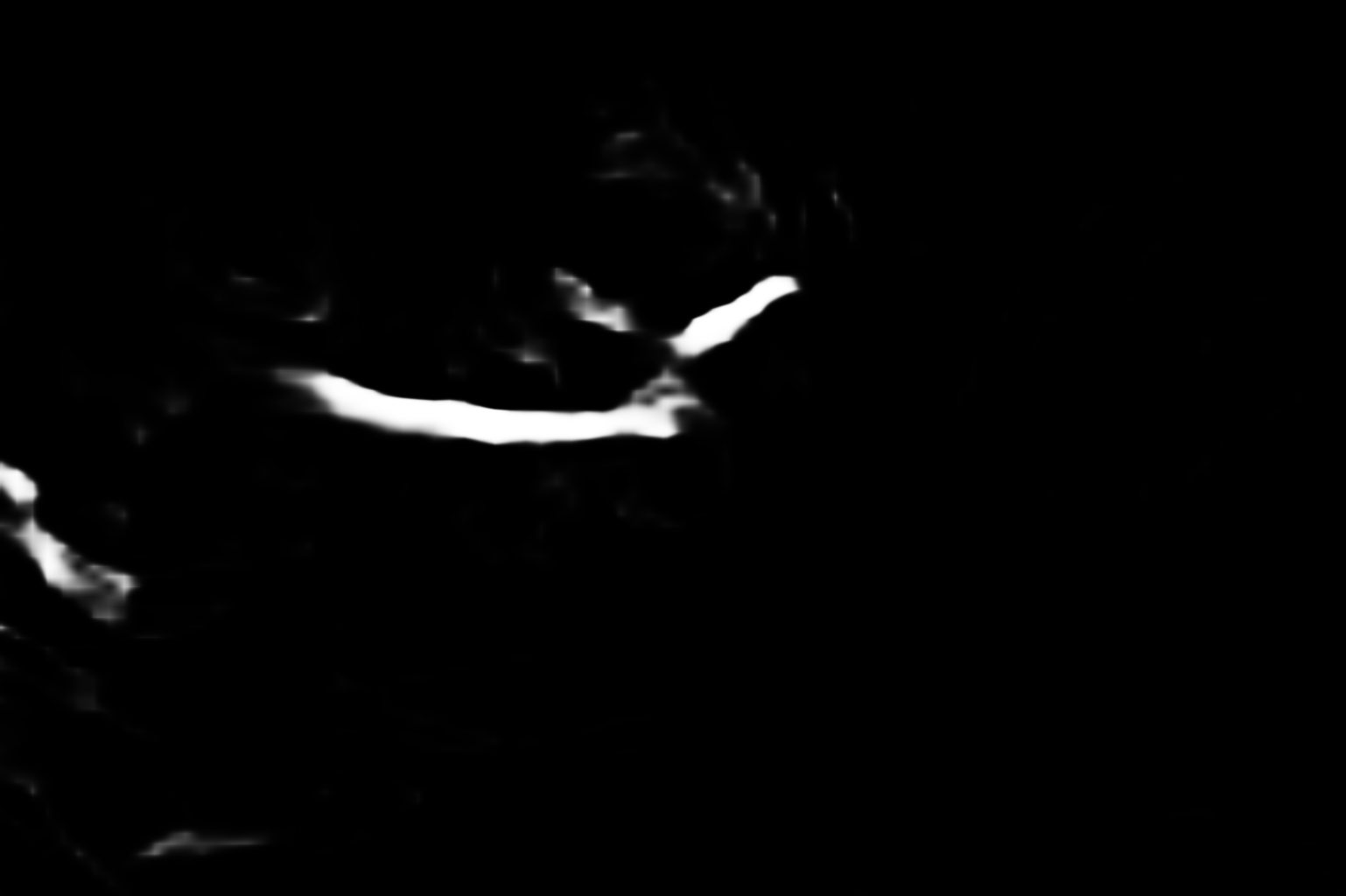}}
  & 
  {\includegraphics[width=0.093\linewidth]{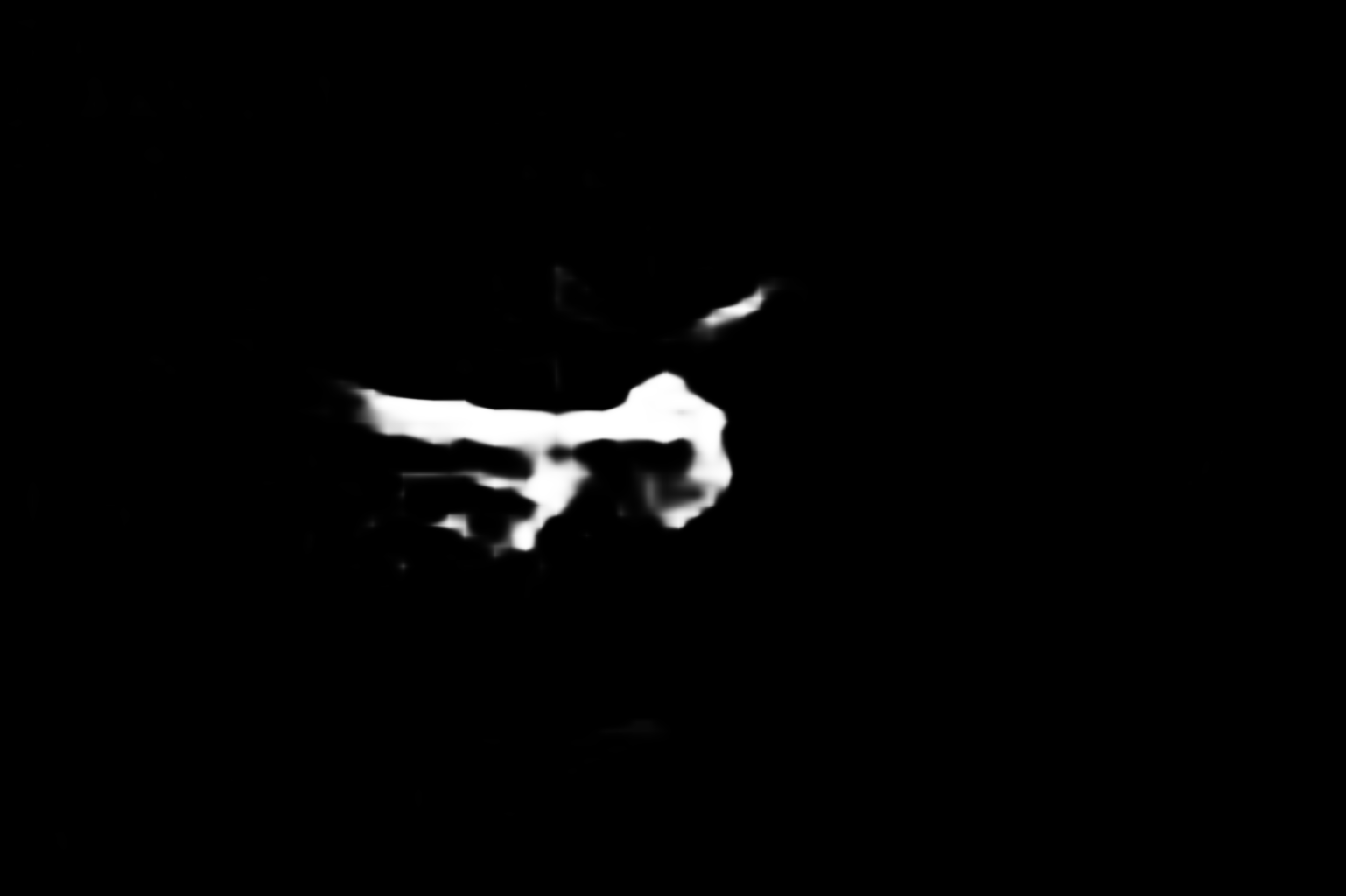}}
  & 
  {\includegraphics[width=0.093\linewidth]{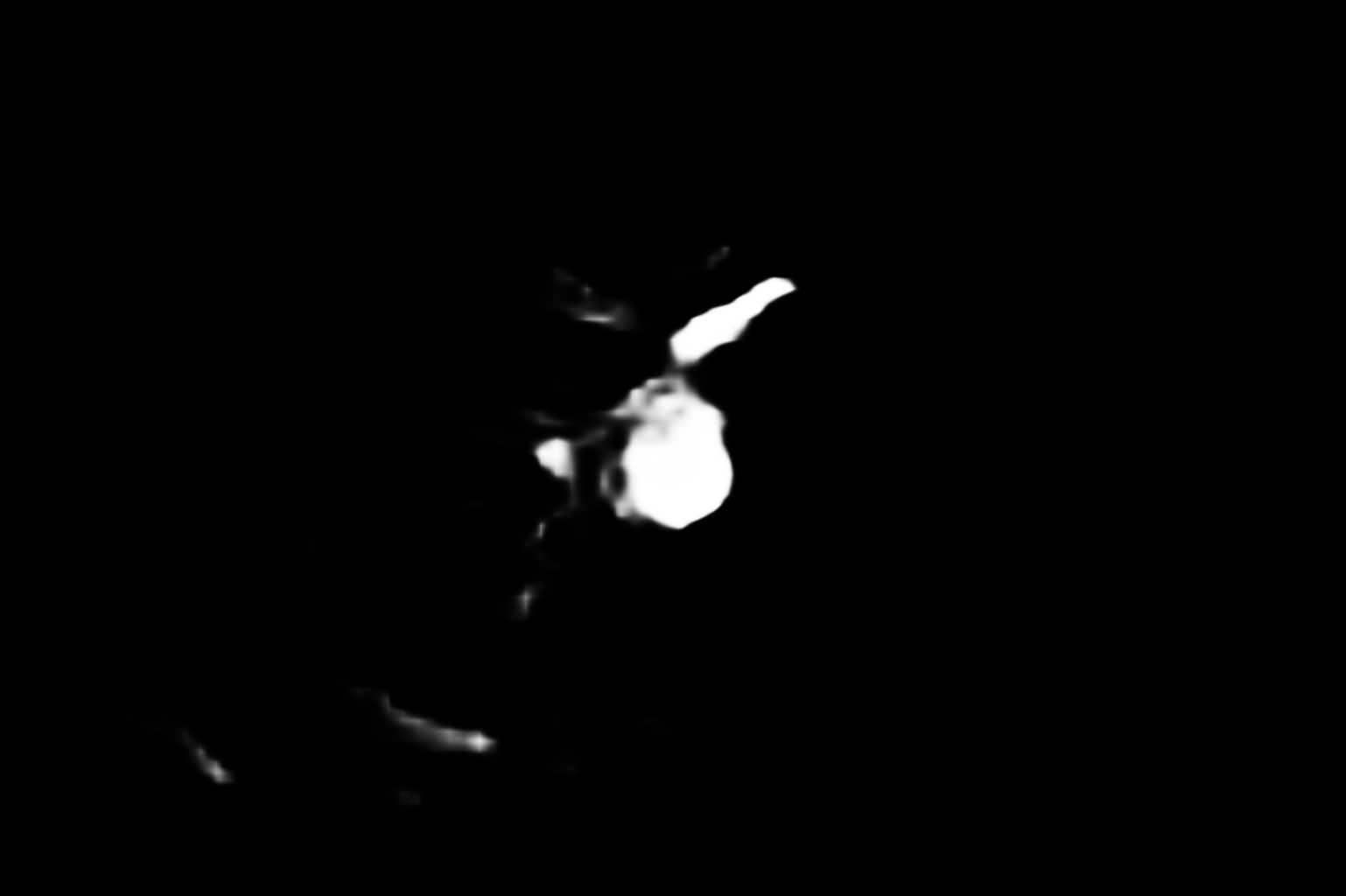}}
  & 
  {\includegraphics[width=0.093\linewidth]{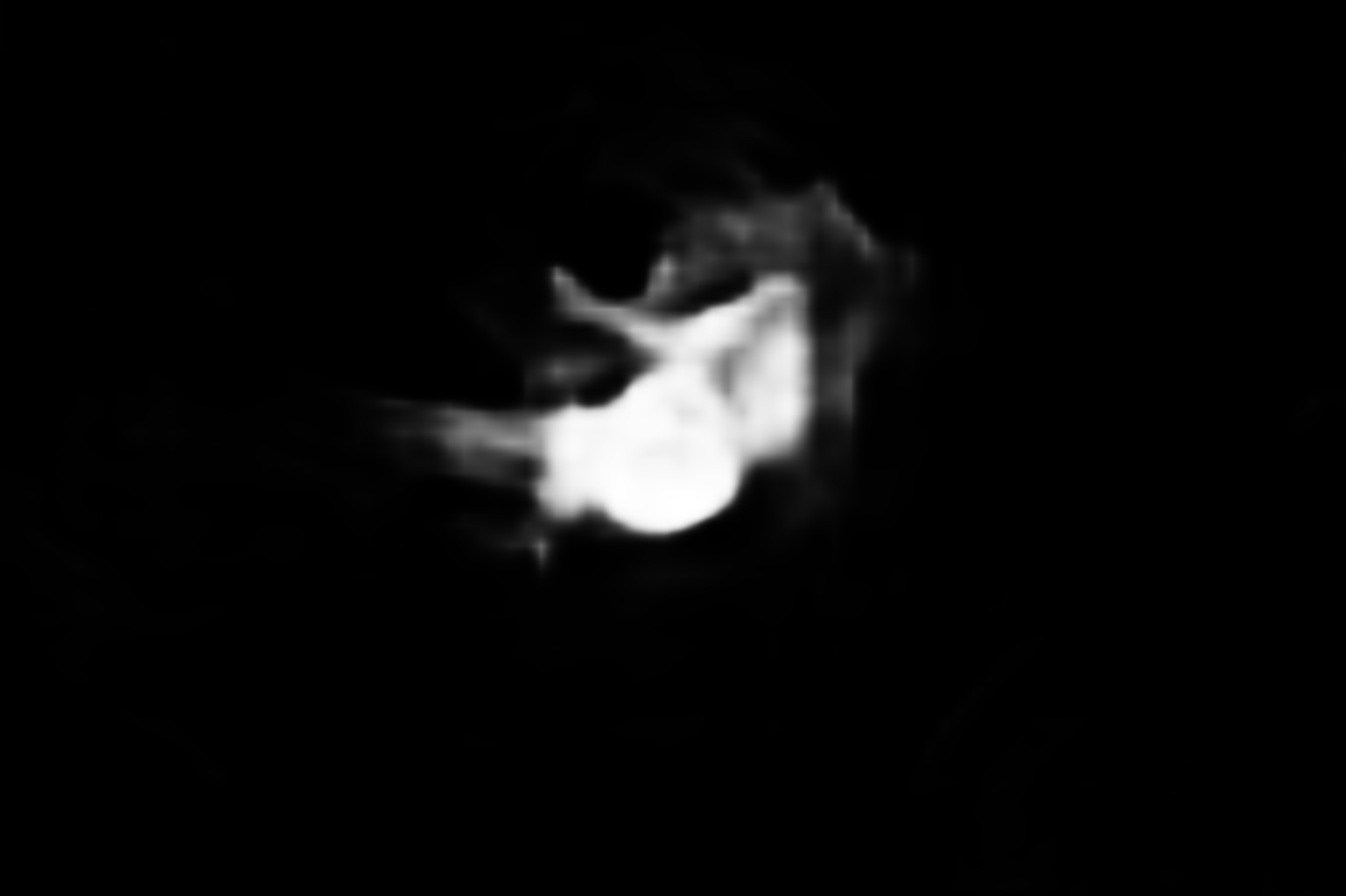}} &
  {\includegraphics[width=0.093\linewidth]{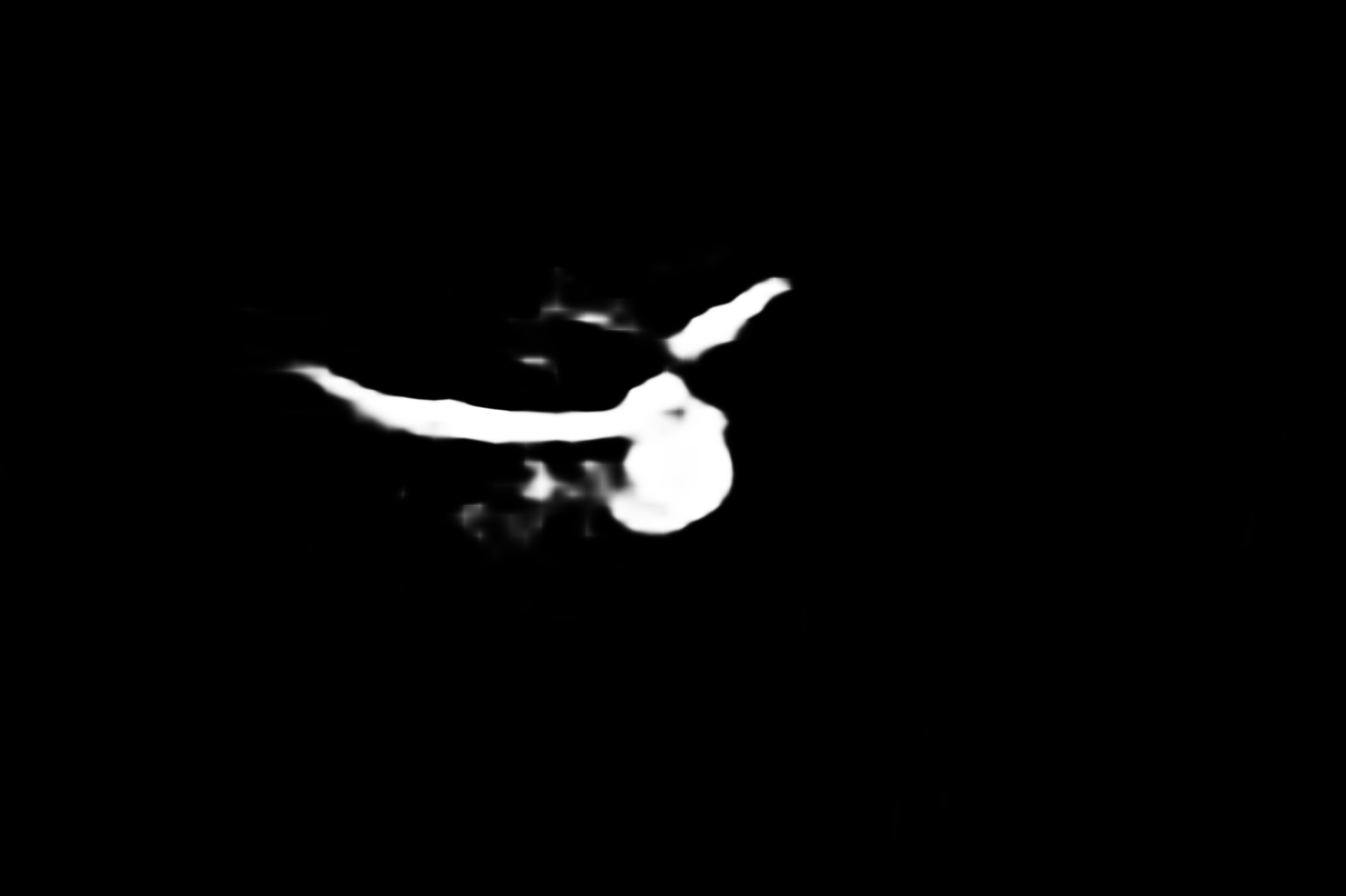}}
  & 
  {\includegraphics[width=0.093\linewidth]{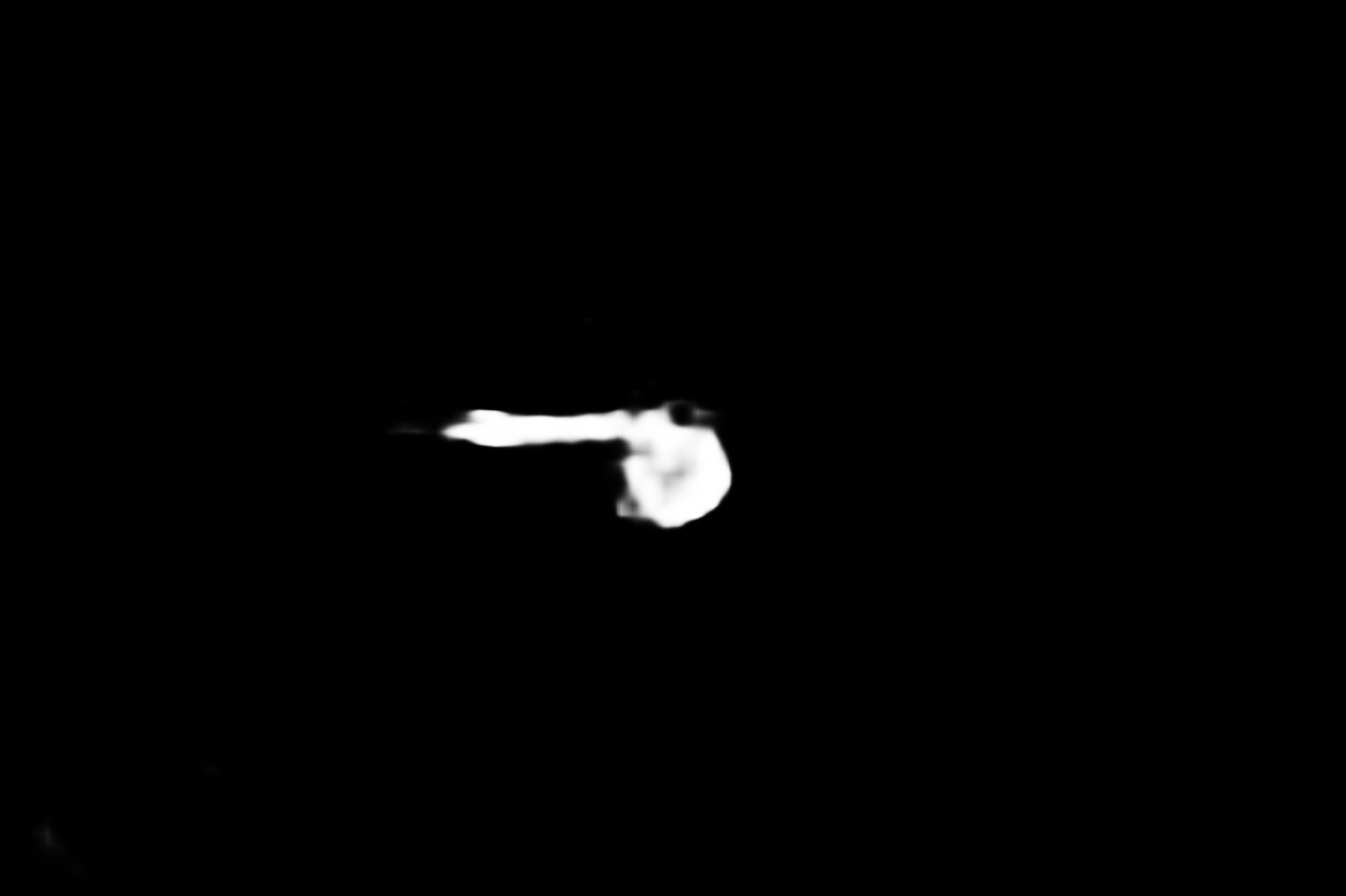}}
  & 
  {\includegraphics[width=0.093\linewidth]{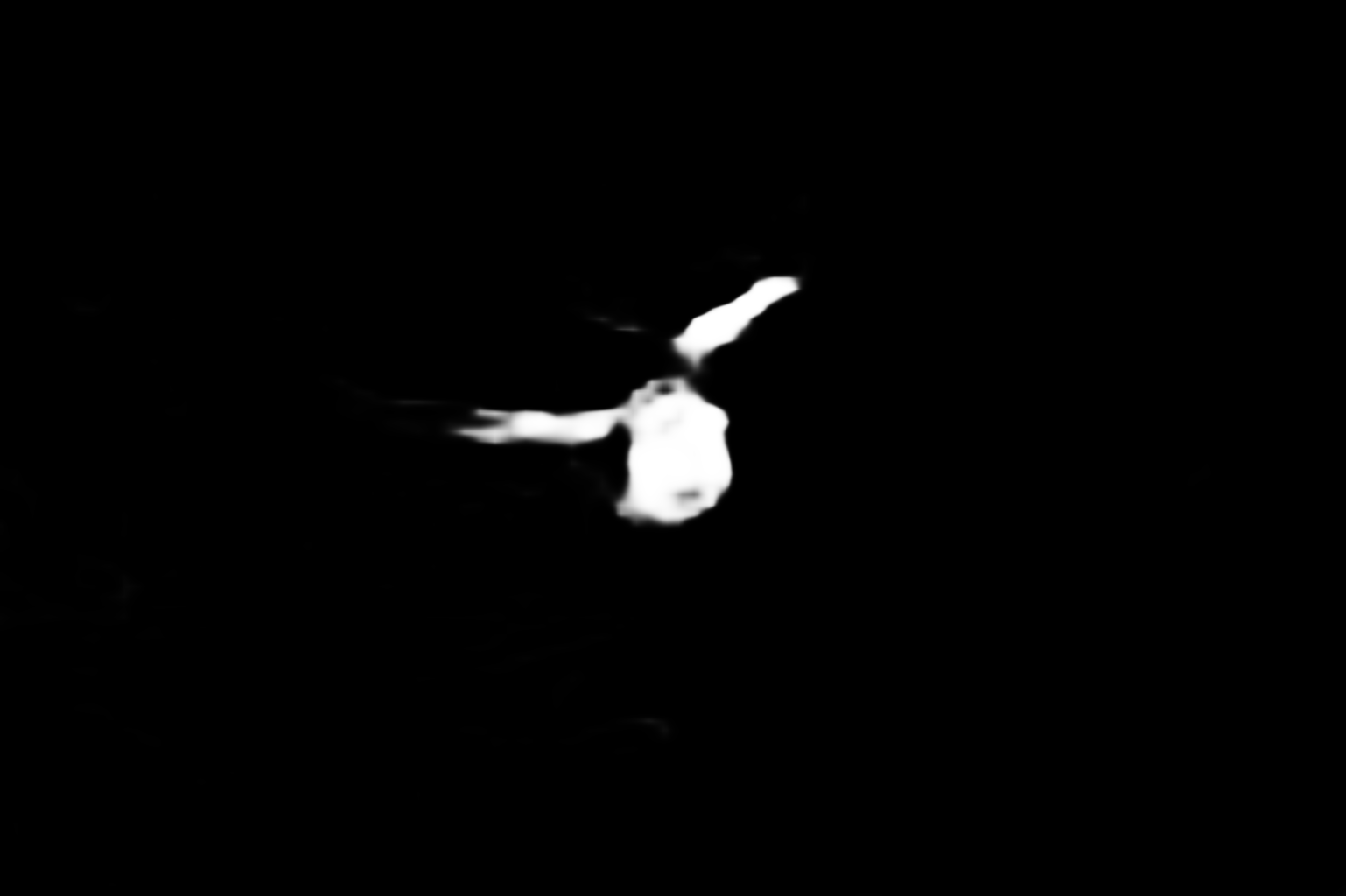}}
  & 
  {\includegraphics[width=0.093\linewidth]{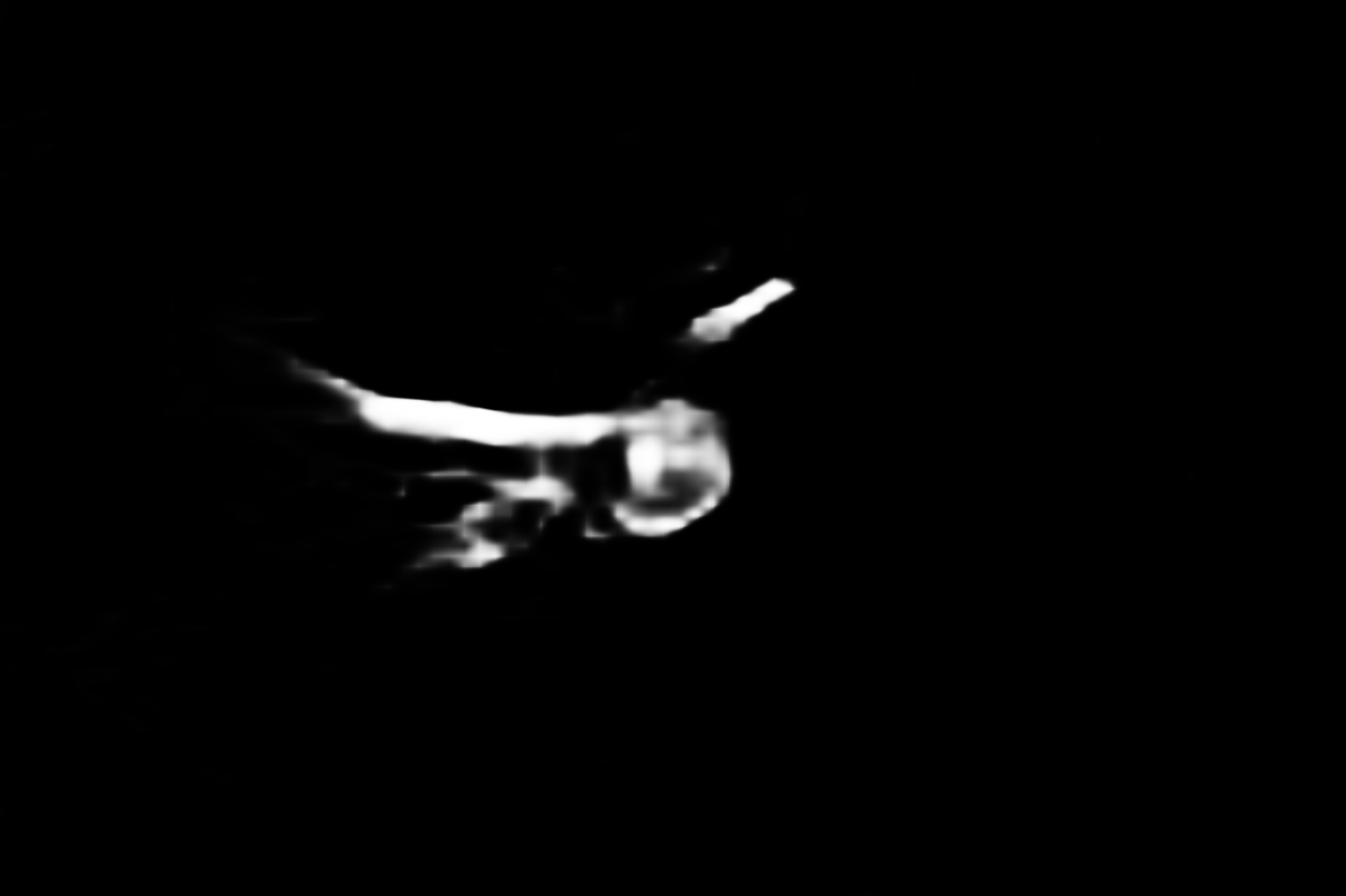}}\\
   \footnotesize{Image}&\footnotesize{GT}&\footnotesize{Concat}&\footnotesize{InvConcat}&\footnotesize{Dot}&\footnotesize{InvDot}&\footnotesize{Gaus}&\footnotesize{InvGaus}&\footnotesize{EGaus}&\footnotesize{InvEGaus}\\
   \end{tabular}
   \end{center}
   \caption{Camouflaged object maps predicted by various attention models.  “Inv-” denotes their corresponding invertible attention counterparts.
   }
\label{fig:predictions_of_attentionmodels}
\end{figure}

\noindent\textbf{Qualitative results:} To further analyse how invertible attention modules affect camouflaged object predictions, we visualize results of the base model as well as those obtained by (invertible) attention models in Fig.~\ref{fig:predictions_of_attentionmodels}.
We observe comparable visual results of the three invertible attention models (Concatenation, Gaussian and Embedded Gaussian) with their corresponding non-invertible form.

\subsection{Discussion}
We observe that the Dot-product attention suffers unstable image reconstruction, degenerate performance in image generation, and significantly decayed performance in camouflaged object detection. It suggests that the response map of Dot-product attention seems unlikely to satisfy the relaxed Lipschitz bound used in our implementation. This is inline with the claim of \cite{kim2020lipschitz} that Dot-product attention may not enjoy a Lipschitz constant. However, all other forms of invertible attentions are functioning well in all experiments.

\section{Conclusion}
\label{sec: conclusion and discussion}
In this paper, we brought global receptive fields to invertible networks by introducing our invertible attention modules. We mathematically proved that a normal attention module can become invertible by constraining its Lipschitz constant. We validated four kinds of invertible attention modules on CIFAR10, SVHN and CelebA datasets with the image reconstruction task. We also embeded our attention module to i-ResNet~\cite{behrmann2019invertible} and shown an improved performance on the image generation task. To demonstrate the expressive power of our invertible attention, we compared its performance on a dense prediction task, namely camouflaged object detection \cite{background_matching,fan2020camouflaged}. Results on the camouflaged object detection task shown that our invertible attention has similar expressive power to the non-invertible ones.

\section{Broader Impact}
\label{sec: broader impact}
The long-term objective of this work is to bring more interpretability into deep learning, for better serving the human society. The short-term impact of this work is to enhance model capacity and capability of invertible networks, and this enhanced invertible image generator could be maliciously used to produce fake profiles. To counter this, researchers could develop fake detectors to recognize the specific pattern in images generated by this model.

\appendix


\section{Supplementary Proof}
We give a more detailed proof for the Theorem 1 here. We start with the $L_1$-norm since it makes the proof simpler. Then we extend the conclusion to the $L_2$-norm and Frobenius norm.


\subsection{Matrix and Vector \texorpdfstring{$L_1$}{L1}-norm}
If $\bm{A}$ is a matrix, then $\bm{A}_j$ will always represent the $j$-th column of $\bm{A}$. For a vector, $\norm{\cdot}_1$ represents its $L_1$-norm, which is equal to the sum of the absolute values of its elements. And for a matrix, $\norm{\cdot}_1$ represents the matrix $L_1$-norm, which is is equal to the maximum of the vector $L_1$-norm of its columns. That is
\begin{equation*}
    \norm{\bm{A}}_1 = \max_j \norm{\bm{A}_j}_1~.
\end{equation*}


\subsection{Main Theorem Proved in \texorpdfstring{$L_1$}{L1}-norm}

\textbf{Lemma 2.1 :} For matrix $\bm{A}$ and vector $\bm{v}$, 
\begin{equation*}
    \norm{\bm{A}\bm{v}}_1 \leq \norm{\bm{A}}_1\norm{\bm{v}}_1~.
\end{equation*}

\textit{Proof. }
\begin{align*}
    \norm{\bm{A}\bm{v}}_1 & = \norm{\sum_j \bm{A}_j\bm{v}_j}_1\\
    & \leq \sum_j \norm{\bm{A}_j\bm{v}_j}_1 &\text{~~~ by triangular inequality}\\
    & = \sum_j \abs{\bm{v}_j} \norm{\bm{A}_j}_1\\
    &= \norm{\bm{A}}_1 \norm{\bm{v}}_1
\end{align*}\\

\textbf{Corollary 2.2 : } For matrix $\bm{G}$ and matrix $\bm{F}$, 
\begin{equation*}
    \norm{\bm{G}\bm{F}}_1 \leq \norm{\bm{G}}_1 \norm{\bm{F}}_1~.
\end{equation*}
This is a well known fact that matrix $L_1$-norm is sub-multiplicative~\cite{characterization_norm}. In fact, in some text books, the terminology matrix norm is only used for those norms that are sub-multiplicative~\cite{rudin1976principles}. \\

\textbf{Theorem 2.3 : }Let $\calX$ be a normed vector space and let $\R^{m\times n}$ represent the vector space of $m\times n$ matrices with $L_1$-norm. Let $\bm{F}: \calX \to \R^{m\times n}$ and $\bm{G}: \calX \to \R ^{m\times m}$ be Lipschitz-continuous functions. Define $\bm{A}: \calX \to R^{m\times n}$ by $\bm{A}(\bm{x}) = \bm{G}(\bm{x})\bm{F}(\bm{x})$ where $\bm{x}\in \calX$~. We further assume the following properties:
\begin{itemize}
    \item $\bm{F}$ has $L_1$-Lipschitz constant $c_F$~.
    \item $\bm{G}$ has $L_1$-Lipschitz constant $c_G$~.
    \item $\norm{\bm{F}(\bm{x})}_1 \leq \mu_F$ for all $\bm{x}\in \calX$~.
    \item $\norm{\bm{G}(\bm{x})}_1 \leq \mu_G$ for all $\bm{x}\in \calX$~.
\end{itemize}
Then $\bm{A}$ has a $L_1$-Lipschitz constant $\mu_Gc_F + \mu_Fc_G$~.\\

\textit{Proof : } For simplicity, denote $\bm{G}(\bm{x}^i)$ and $\bm{F}(\bm{x}^i)$ by $\bm{G}_i$ and $\bm{F}_i$ respectively. Now

    

\begin{align*}
    \norm{\bm{A}(\bm{x}^1) - \bm{A}(\bm{x}^2)}_1 &= \norm{\bm{G}_1\bm{F}_1 - \bm{G}_2\bm{F}_2}_1\\
    &= \norm{\bm{G}_1(\bm{F}_1 - \bm{F}_2) + (\bm{G}_1 - \bm{G}_2)\bm{F}_2}_1\\
    & \leq \norm{\bm{G}_1(\bm{F}_1 - \bm{F}_2)}_1 + \norm{(\bm{G}_1 - \bm{G}_2)\bm{F}_2}_1 & \text{~~~ by triangular inequality}\\
    & \leq \norm{\bm{G}_1}_1\norm{\bm{F}_1 - \bm{F}_2}_1 + \norm{\bm{G}_1 - \bm{G}_2}_1\norm{\bm{F}_2}_1 &\text{~~~ by Corollary 2.2}\\
    & \leq \mu_G c_F \norm{\bm{x}^1 - \bm{x}^2}_1 + c_G \norm{\bm{x}^1 - \bm{x}^2}_1\mu_F~,
\end{align*}
which gives the requested result.\\

\textbf{Another way of understanding : } Lipschitz constant is highly related to derivative. Assume $\bm{F}, \bm{G}$ are both differentiable, then use the product rule of derivative, we can write the derivative of $\bm{A}$
\begin{equation*}
    \bm{A}' = \bm{F}'\bm{G} + \bm{F}\bm{G}'~.
\end{equation*}
So, if $\bm{A}'$ needs to be bounded, then $\bm{F}, \bm{G}, \bm{F}', \bm{G}'$ all need to be bounded. The bounds on $\bm{F}, \bm{G}$ are $\mu_F, \mu_G$ whereas the bounds on $\bm{F}', \bm{G}'$ are Lipschitz constant $c_F, c_G$. 


\subsection{Extend to Other Norms}
\label{sec: supp norm extension}
The key step in the proof of Theorem 2.3 is Corollary 2.2 (matrix $L_1$-norm is sub-multiplicative), so the proof holds equally well for any norm that satisfies Corollary 2.2. We prove $L_2$-norm and Frobenius-norm are sub-multiplicative here, and then the theorem and its proof could hold valid for $L_2$-norm and Frobenius-norm also.

\subsubsection{\texorpdfstring{$L_2$}{L2}-norm}
For a vector $\bm{v}$, the $L_2$-norm is defined to be
\begin{equation*}
    \norm{\bm{v}}_2 = \sqrt{\sum_i \bm{v}_i^2}~.
\end{equation*}
And for a matrix $\bm{A}$, the $L_2$-norm is 
\begin{equation*}
    \norm{\bm{A}}_2 = \max_{\bm{v} \neq \mathbf{0}} \frac{\norm{\bm{A}\bm{v}}_2}{\norm{\bm{v}}_2}~.
\end{equation*}
We now prove that $L_2$-norm is sub-multiplicative.

\textbf{Lemma 3.1 : } For any matrix $\bm{A}$ and any vector $\bm{v}$, $\norm{\bm{A}\bm{v}}_2 \leq \norm{\bm{A}}_2\norm{\bm{v}}_2$~.

\textit{Proof : }

For $\bm{v}\neq \mathbf{0}$, 
\begin{equation*}
    \norm{\bm{A}\bm{v}}_2 = \frac{\norm{\bm{A}\bm{v}}_2}{\norm{\bm{v}}_2} \norm{\bm{v}}_2 \leq \left(\max_{\bm{w}\neq \mathbf{0}}\frac{\norm{\bm{A}\bm{w}}_2}{\norm{\bm{w}}_2}\right)\norm{\bm{v}}_2 = \norm{A}_2\norm{v}_2~.
\end{equation*}

For $\bm{v} = \mathbf{0}$, 
\begin{equation*}
    \norm{\bm{A}\bm{v}}_2 = \norm{\bm{A}\mathbf{0}}_2 = 0 = \norm{A}_2\norm{\mathbf{0}}_2 = \norm{\bm{A}}_2 \norm{\bm{v}}_2~.
\end{equation*}

Therefore, $\norm{\bm{A}\bm{v}}_2 \leq \norm{\bm{A}}_2\norm{\bm{v}}_2$ always holds. \\

\textbf{Corollary 3.2 : } $L_2$-norm is sub-multiplicative. For any two matrices $\bm{G}$ and $\bm{F}$, 
\begin{equation*}
    \norm{\bm{G}\bm{F}}_2 \leq \norm{\bm{G}}_2\norm{\bm{F}}_2~.
\end{equation*}

\textit{Proof : }

\begin{align*}
    \norm{\bm{G}\bm{F}}_2 & = \max_{\bm{v}\neq \mathbf{0}} \frac{\norm{\bm{G}\bm{F}\bm{v}}_2}{\norm{\bm{v}}_2} &\text{~~~ by the definition of $L_2$-norm}\\
    &= \frac{\norm{\bm{G}\bm{F}\bm{v}'}_2}{\norm{\bm{v}'}_2} &\text{~~~ assume $\frac{\norm{\bm{G}\bm{F}\bm{v}'}_2}{\norm{\bm{v}'}_2} = \max_{\bm{v}\neq \mathbf{0}} \frac{\norm{\bm{G}\bm{F}\bm{v}}_2}{\norm{\bm{v}}_2}$}\\
    & \leq \frac{\norm{\bm{G}}_2\norm{\bm{F}\bm{v}'}_2}{\norm{\bm{v}'}_2} & \text{~~~ by Lemma 3.1}\\
    & \leq \norm{\bm{G}}_2 \left(\max_{\bm{v}' \neq \mathbf{0}} \frac{\norm{\bm{F}\bm{v}'}_2}{\norm{\bm{v}'}} \right)\\
    &\leq \norm{G}_2\norm{F}_2~, &\text{~~~  by the definition of $L_2$-norm}~,
\end{align*}
which proved that $L_2$-norm is sub-multiplicative.

\subsubsection{Frobenius-norm}
Frobenius-norm is specifically defined for a matrix, 
\begin{equation*}
    \norm{\bm{A}}_F = \sqrt{\sum_{ij}\bm{A}_{ij}^2}~.
\end{equation*}

\textbf{Lemma 3.2 :} For any matrix $\bm{A}$, its $L_2$-norm is no larger than its Frobenius-norm,
\begin{equation*}
    \norm{\bm{A}}_2 \leq \norm{\bm{A}}_F~.
\end{equation*}\\

\textit{Proof : }
\begin{align*}
    \norm{\bm{A}}_2^2 & = \left(\max_{\bm{v}\neq \mathbf{0}} \frac{\norm{\bm{A}\bm{v}}_2}{\norm{\bm{v}}_2}\right)^2 = \left( \max_{\norm{\bm{v}}_2 = 1} \norm{\bm{A}\bm{v}}_2 \right)^2\\
    & = \norm{ \bm{A}\bm{v}^\star }_2^2 &\text{~~~assume $\norm{\bm{v}^\star}_2=1$ and $\norm{ \bm{A}\bm{v}^\star }_2 = \max_{\bm{v}\neq \mathbf{0}} \frac{\norm{\bm{A}\bm{v}}_2}{\norm{\bm{v}}_2}$}\\
    & = \norm{\sum_j \bm{v}^\star_j \bm{A}_j }_2^2 \\
    &\leq \left( \sum_j \norm{\bm{v}^\star_j \bm{A}_j}_2 \right)^2 =\left( \sum_j \abs{\bm{v}^\star_j} \norm{\bm{A}_j}_2 \right)^2&\text{~~~ by triangular inequality} \\
    & \leq \left( \sum_j (\bm{v}^\star_j)^2 \right)\left( \sum_j \norm{\bm{A}_j}_2^2 \right) &\text{~~~ by Cauchy-Schwartz inequality}\\
    & = \norm{\bm{v}^\star}_2^2 \norm{\bm{A}}_F^2 = \norm{\bm{A}}_F^2
\end{align*}\\

\textbf{Corollary 3.3 : } Frobenius-norm is sub-multiplicative. For any two matrices $\bm{G}, \bm{F}$,
\begin{equation*}
    \norm{\bm{G}\bm{F}}_F \leq \norm{\bm{G}}_F\norm{\bm{F}}_F~.
\end{equation*}

\textit{Proof : }
The squared Frobenius-norm could be decomposed as the sum of vector $L_2$-norms of its columns. Thus,
\begin{align*}
    \norm{\bm{G}\bm{F}}_F^2 &= \sum_j \norm{\bm{G}\bm{F}_j}_2^2 & \text{~~~ by definition of Frobenius-norm and $L_2$-norm}\\
    &\leq \sum_j \norm{\bm{G}}_2^2 \norm{\bm{F}_j}_2^2 & \text{~~~ by Lemma 3.1}\\
    &= \norm{\bm{G}}_2^2 \sum_j \norm{\bm{F}_j}_2^2 \\
    &= \norm{\bm{G}}_2^2 \norm{\bm{F}}_F^2 &\text{~~~ by definition of Frobenius-norm and $L_2$-norm}\\
    &\leq \norm{\bm{G}}_F^2 \norm{\bm{F}}_F^2 &\text{by Lemma 3.2}
\end{align*}

\section{Constraining the Lipshchitz constant in PyTorch}
We use the pre-forwarding hook in PyTorch to enforce the Lipshchitz bound on convolution. The basic idea of such mechanism is that, before the actual forward pass, the hook is invoked to check and prune the weight parameters of this convolution, making sure the bound holds valid during the actual forward pass. The problem is, this hook cannot guarantee the bound after the weight update.

An iteration in a typical training process includes a forward pass, a gradient back-propagation and a weight update. As the weight parameters are not the same as the output of the hook after weight update, there is no strict guarantee that the Lipschitz bound will still hold. 
The ideal moment to invoke this hook is after the weight update, but there has not been such an interface in PyTorch. However, we empirically find that using the pre-forwarding hook works well in most cases and if it fails, we will handle it using the techniques below.




\section{Enhancing Invertibility}
As we mentioned in the main manuscript, the relaxed conditions are not aiming to make attention strictly invertible but to achieve a balance between the invertibility and the expressive power. It means that in theory, our model may fail in some cases due to the relaxed settings and the model may run into an unstable state due to the problem of pre-forwarding hook as described in previous section. We use the following techniques to avoid these issues.

The following techniques could enforce a stronger invertibility by applying a stronger Lipschitz bound over the residual branch. We rank them according to their tightness on that bound. We wouldn't recommend using a much stronger trick if a less effective one could already solve the problem, because it will significantly reduce the capacity of the model.

\begin{enumerate}
    \item Shrink learning rate.
    \item Use activation functions with continuous derivative. For example, use ELU but not ReLU.
    \item Shrink the user-specified Lipschitz constant $c$, and enlarge the inverse iteration number $N$. A detailed discussion over these two parameters can be found in \cite{behrmann2019invertible}.
    \item Shrink the column sum of the response map. 
    \item Apply the same Lipschitz constraint to the linear transform (if there is) in the computation of response function. 
    \item Change from each column of the
    response map must sum to $1$ to all elements of the response map must sum to $1$.
\end{enumerate}

\section{Numeric Issues of Gaussian Type Attention}
\label{sec: supp gaussian issue}
There could be a numeric overflow issue for Gaussian and Embedded Gaussian type attention, whether invertible or non-invertible. Taking Gaussian attention as an example, its response function
\begin{equation*}
    \bm{m}(\bm{x}_i, \bm{x}_j) = e^{\bm{x}_i^\top \bm{x}_j}
\end{equation*}
will exponentially increase when $\bm{x}_{i}^\top \bm{x}_j$ gets larger. And for a typical convolution network with a large number of channels, $\bm{x}_{i}^\top \bm{x}_j$ usually is a large value. This may cause floating-point overflow problem for the normalizing coefficient, which is 
\begin{equation*}
    \mathcal{C}(\bm{x}) = \left\{ 
    \begin{array}{ll}
        \sum_{\forall j}\bm{m}(\bm{x}_i, \bm{x}_j) & \text{non-invertible} \\
        \sum_{\forall i}\bm{m}(\bm{x}_i, \bm{x}_j) & \text{invertible}
    \end{array}~.
    \right.
\end{equation*}
If this overflow happens, the training loss will immediately become \texttt{NaN} and the model is not trainable any more. A potential idea is to do maximum suppression on the response function
\begin{equation*}
    \bm{m}(\bm{x}_i, \bm{x}_j) = \left\{ 
    \begin{array}{ll}
        e^{\bm{x}_i^\top \bm{x}_j} - \max_j\left(e^{\bm{x}_i^\top \bm{x}_j}\right)   & \text{non-invertible} \\
        e^{\bm{x}_i^\top \bm{x}_j} - \max_i\left(e^{\bm{x}_i^\top \bm{x}_j}\right) & \text{invertible}
    \end{array}~.
    \right. 
\end{equation*}
Note that the Gaussian and Embedded Gaussian attention now do not satisfy the first condition in Section 2.4 of the main manuscript any more. A typical operation like wrapping them with another activation function is not feasible here as all response values are smaller than or equal to 0 after maximum suppression, which would only produce 0 if wrapped with normal activation functions like ReLU. However, this problem arises from the definition of Gaussian, irrelevant to our focus on invertibility in this work. 

\section{Fast Computation of Log-determinant}
\label{supp: fast log det}
\subsection{Why Log-determinant}
Normalizing Flows follow a specific mathematical requirement. Suppose $\bm{z}\in \mathbb{R}^d$ and follows a pre-defined distribution $\bm{z}\sim p_{\bm{z}}(\bm{z})$, we also need to have a bijective transformation $\bm{\Phi}: \mathbb{R}^d \to \mathbb{R}^d$ that will do $\bm{x} = \bm{\Phi}(\bm{z})$. Define $\bm{F} = \bm{\Phi}^{-1}$, we then can compute the likelihood of any $\bm{x}$ by change of variable formula
\begin{equation*}
    \label{eq: 32}
    \ln p_{\bm{x}}(\bm{x}) = \ln p_{\bm{z}}(\bm{z}) + \ln|\det \bm{J}_{\bm{F}}(\bm{x})|~,
\end{equation*}
where $\bm{J}_{\bm{F}}(\bm{x})$ is the Jacobian of $\bm{F}$ evaluated at $\bm{x}$. Models of this form are known as Normalizing Flows~\cite{rezende15}. And the typical training objective of such models is to make the right-hand-side of the equation above as large as possible.

\subsection{Fast Estimation of Log-determinant}
\label{sec: supp fast estimation}
i-ResNet~\cite{behrmann2019invertible} proposed a fast estimation algorithm for residual structures, and it could be directly applied on our invertible attention module as our module is also a residual structure. 

The Lipschitz constraint on the residual branch $\bm{g}(\bm{x})$ yields a positive determinant for the whole block $\bm{F}(\bm{x}) = \bm{x} + \bm{g}(\bm{x})$. Therefore,
\[|\det \bm{J}_{\bm{F}}| = \det \bm{J}_{\bm{F}}(\bm{x})~.\]
With matrix identity~\cite{Matrix-Identity}, we get
\[  \ln |\det \bm{J}_{\bm{F}}(\bm{x})|  = \ln\det\bm{J}_{\bm{F}}(\bm{x}) = \tr(\ln \bm{J}_{\bm{F}})~. \]
Consider $\bm{F} = \bm{I} + \bm{g}(\bm{x})$, 
\begin{equation*}
    \label{eq: 33}
    \ln |\det \bm{J}_{\bm{F}}(\bm{x})| = \tr(\ln (\bm{I} + \bm{J}_{\bm{g}}(\bm{x})))~.
\end{equation*}
Substitute this result into the objective function of Normalizing Flows,  
\begin{equation*}
    \label{eq: 34}
    \ln p_{\bm{x}}(\bm{x}) = \ln p_{\bm{z}}(\bm{z}) + \tr (\ln (\bm{I} + \bm{J}_{\bm{g}}(\bm{x})))~.
\end{equation*}
When $\norm{\bm{J}_{\bm{g}}}_2 < 1$, $\tr(\ln(\bm{I} + \bm{J}_{\bm{g}}(\bm{x})))$ can be expressed as sum of a power series 
\begin{equation*}
    \label{eq: 35}
    \tr(\ln(\bm{I} + \bm{J}_{\bm{g}}(\bm{x}))) = \sum_{k=1}^{\infty}(-1)^{k+1}\frac{\tr(\bm{J}_{\bm{g}}^k)}{k}~.
\end{equation*}
Up to this point, we have already shifted our focus from computing the Jacobian of the whole block to the Jacobian of residual branch only. And this Jacobian matrix is actually computed automatically by the differentiation engine of PyTorch. 

But this computation is conducted internally and concurrently by PyTorch and the full Jacobian matrix is not directly accessible. This design of PyTorch is for saving video memory and accelerating computation.

For us, we need to compute the trace of that Jacobian matrix. We then turn to Hutchinson trace estimator~\cite{Hutchinson}. Vector $\bm{v} \in\mathbb{R}^N$ is randomly sampled from a distribution that satisfies $\mathrm{E}(\bm{v}) = 0$ and $\mathrm{Cov}(v) = \bm{I}$, and the trace of a $N\times N$ matrix $\bm{J}$ can then be estimated by 
\begin{equation*}
    \label{eq: 36}
    \tr(\bm{J}) = \mathrm{E}_{p(\bm{v})}[\bm{v}^\top\bm{J}\bm{v}]~.
\end{equation*}
We also need to compute the trace of that Jacobian matrix raised to some exponential. It could be done by
\begin{equation*}
    \label{eq: 37}
    \tr(\bm{J}^k) = \mathrm{E}_{p(\bm{v})}[\bm{v}^\top\underbrace{\bm{J}\bm{J}...\bm{J}}_\text{k}\bm{v}]~.
\end{equation*}
We can now efficiently compute the log-determinant of our invertible attention module.

\section{Supplementary Experiments}
\subsection{Validating Invertibility}
\label{sec: supp valid inv}
We use the same metrics as in the main manuscript, \ie MSE, SSIM, and V-score,  to measure the reconstruction quality on $218\times 178$ CelebA images in Table \ref{tab: quanti recon}.

\begin{table}[h]
    \centering
    \caption{Measure the reconstruction quality on $218\times 178$ CelebA images.}
    \begin{tabular}{lccc}
    \toprule
                & MSE           & SSIM  & V-score  \\
    \midrule
Gaussian        & 64.975        & 0.991 & 87.000\% \\
Embed. Gaussian & 27.506        & 0.994 & 93.900\% \\
Dot-product     & 251.231       & 0.976 & 94.100\% \\
Concatenation   & $1.555e^{-6}$ & 1.000 & 100\%\\
    \bottomrule
    \end{tabular}
    \label{tab: quanti recon}
\end{table}

We also provide the full implementation code for this experiment\footnote{https://github.com/Schwartz-Zha/InvertibleAttention}.

\subsection{Generative Modelling}
\label{sec: supp generative modelling}
We provide the full implementation code for this experiment as well. Note that this includes our fast estimation algorithm mentioned in Appendix \ref{supp: fast log det}.

\subsection{Invertible Attention \emph{vs.} Non-invertible Attention in Discriminative Learning}
\label{sec: supp discriminative}

We introduce the detailed implementation about the Non-invertible Attention in Discriminative Learning for camouflaged object detection.

\noindent\textbf{Dataset:}
We train the models using the COD10K training dataset\cite{fan2020camouflaged} of size 4,040, and test them on four test datasets:  CAMO \cite{le2019anabranch} (250 images), CHAMELEON \cite{Chameleon2018} (76 images), the COD10K testing dataset (2,026 images), and NC4K\cite{yunqiu_cod21} (4,121 images).

\noindent\textbf{Evaluation metrics:}
We use four evaluation indicators to measure model performance, including Mean Absolute Error $\mathcal{M}$, Mean F-measure ($F_{\beta}$), Mean E-measure ($E_{\xi}$) \cite{fan2018enhanced} and S-measure ($S_{\alpha}$) \cite{fan2017structure}.

\textbf{MAE} $\mathcal{M}$ is defined as the pixel-wise difference between the predicted $c$ and the pixel-wise binary ground-truth $y$:
\begin{equation*}
    \begin{aligned}
    \text{MAE} = \frac{1}{H\times W}|c-y|,
    \end{aligned}
\end{equation*}
where $H$ and $W$ are the height and width of $c$ correspondingly.

\textbf{F-measure} $F_{\beta}$ is a region based similarity metric, and we provide the mean F-measure using varying fixed (0-255) thresholds.

\textbf{E-measure} $E_{\xi}$ is the recent proposed Enhanced alignment
measure~\cite{fan2018enhanced} in the binary map evaluation field to jointly capture image-level statistics and local pixel matching information.

\textbf{S-measure} $S_{\alpha}$ is a structure based measure ~\cite{fan2017structure}, which combines the region-aware ($S_r$) and object-aware ($S_o$) structural similarity as their final structure metric:
\begin{equation*}
\label{equ:S-measure}
S_{\alpha} = \alpha S_o+(1-\alpha) S_r,
\end{equation*}
where $\alpha\!\in\![0,1]$ is the balance parameter and set to 0.5 as default.

\begin{figure*}[tp]
   \begin{center}
   \begin{tabular}{c@{ }}
   {\includegraphics[width=0.85\linewidth]{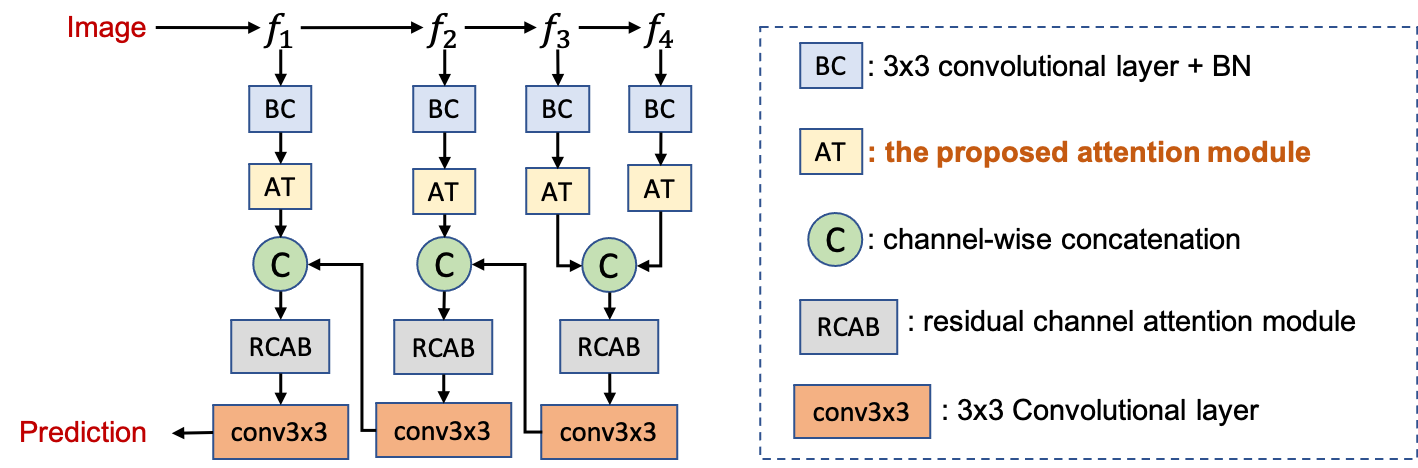}} \\
   \end{tabular}
   \end{center}
   \caption{\footnotesize{The camouflaged object detection network.}
   }
\label{fig:cod_model}
\end{figure*}

\noindent\textbf{Network structure:}
We show structure of the camouflaged object detection network in Fig.~\ref{fig:cod_model}, where \enquote{AT} represents our proposed attention modules in Table 4 of the main paper, and \enquote{RCAB} is the residual channel attention module from \cite{rca_eccv}. We use binary cross-entropy as our loss function.

\noindent\textbf{Training details:}
We train our model in Pytorch with the ResNet50 \cite{he2015deep} trained on ImageNet-1K \cite{imagenet_1k},
and other newly added layers are randomly initialized. We resize all the images and ground truth to $352\times352$. The maximum epoch is 30. The initial learning rates are $2.5 \times 10^{-5}$ for all the related models. The whole training takes 5 hours on average for each related model with batch size 6 on one NVIDIA GTX 2080Ti GPUs for all the models.

\bibliographystyle{plain}
\bibliography{main.bib}


\end{document}